%% file: main.tex
\documentclass[letterpaper]{article}
\usepackage{proceed2e}
\usepackage[margin=1in]{geometry}

\usepackage{times}

\usepackage[labelfont=bf]{caption}
\usepackage[mathscr]{eucal}
\usepackage[noend]{algpseudocode}
\usepackage[round]{natbib}
\usepackage[subrefformat=parens]{subcaption}
\usepackage[T1]{fontenc}
\usepackage[title,page]{appendix}
\usepackage[usenames,dvipsnames,svgnames]{xcolor}

\usepackage{algorithmicx}
\usepackage{algorithm}
\usepackage{amsfonts}
\usepackage{amsmath}
\usepackage{amssymb}
\usepackage{amsthm}
\usepackage{array}
\usepackage{bm}
\usepackage{booktabs}
\usepackage{colortbl}
\usepackage{enumitem}
\usepackage{fancyvrb}
\usepackage{framed}
\usepackage{graphicx}
\usepackage{lipsum}
\usepackage{listings}
\usepackage{mdframed}
\usepackage{multirow}
\usepackage{placeins}
\usepackage{standalone}
\usepackage{tabularx}
\usepackage{textcomp}
\usepackage{tikz}
\usepackage{url}
\usepackage{varwidth}

\usetikzlibrary{bayesnet}
\usetikzlibrary{calc}
\usetikzlibrary{decorations.pathreplacing}
\usetikzlibrary{patterns}
\usetikzlibrary{positioning}

\input{commands}

\title{Probabilistic Search for Structured Data via\\
Probabilistic Programming and Nonparametric Bayes}

\author{
    Feras Saad,\; Leonardo Casarsa,\; and Vikash Mansinghka \\
    Probabilistic Computing Project\\
    Massachusetts Institute of Technology
}

\begin{document}

\maketitle

\input{abstract}

\input{introduction}

\input{bayesian-relevance}

\input{crosscat-relevance}

\input{applications}

\input{discussion}

\clearpage

\input{acknowledgements}

\bibliographystyle{plainnat}
\bibliography{main}

\clearpage

\appendixtitleoff

\begin{appendices}

\input{bayesdb-integration}

\clearpage

\input{figures/cosine}

\clearpage

\input{figures/baselines}

\clearpage

\input{figures/cars}

\end{appendices}

\end{document}

%% file: commands.tex
\newcommand{\bphi}{\bm\phi}

\newcommand{\mtC}{\mathcal{C}}

\newcommand{\mtI}{\mathcal{I}}

\newcommand{\mtQ}{\mathcal{Q}}
\newcommand{\mtR}{\mathcal{R}}

\newcommand{\mtV}{\mathcal{V}}

\newcommand{\D}{\mathscr{D}}

\newcommand{\X}{\mathbf{X}}
\newcommand{\bv}{\mathbf{v}}

\newcommand{\x}{\mathbf{x}}

\newcommand{\z}{\mathbf{z}}

\newcommand{\set}[1]{\left\{{#1}\right\}}

\renewcommand{\ttdefault}{cmtt}

\algblockdefx[ELSETWO]{ElseTwo}{EndElseTwo}
  {\textbf{else}}{\textbf{end else}}
\makeatletter
\ifthenelse{\equal{\ALG@noend}{t}}%
  {\algtext*{EndElseTwo}}
  {}%
\makeatother

\newcolumntype{Y}{>{\centering\arraybackslash}X}

%% file: abstract.tex

\begin{abstract}
Databases are widespread, yet extracting relevant data can be difficult.
Without substantial domain knowledge, multivariate search queries often
return sparse or uninformative results.
This paper introduces an approach for searching structured data based
on probabilistic programming and nonparametric Bayes.
Users specify queries in a probabilistic language that combines
standard SQL database search operators with an information theoretic
ranking function called {\em predictive relevance}.
Predictive relevance can be calculated by a fast sparse matrix algorithm based
on posterior samples from CrossCat, a nonparametric Bayesian model for
high-dimensional, heterogeneously-typed data tables.
The result is a flexible search technique that applies to a broad class of
information retrieval problems, which we integrate into BayesDB, a probabilistic
programming platform for probabilistic data analysis.
This paper demonstrates applications to databases of US colleges, global
macroeconomic indicators of public health, and classic cars.
We found that human evaluators often prefer the results from probabilistic
search to results from a standard baseline.
\end{abstract}

%% file: introduction.tex

\section{Introduction}
\label{sec:introduction}

We are surrounded by multivariate data, yet it is difficult to search.
Consider the problem of finding a university with a city campus, low student
debt, high investment in student instruction, and tuition fees within a certain
budget.
The US College Scorecard dataset
\citep{council2015} contains these variables plus hundreds of others.
However, choosing thresholds for the quantitative variables --- debt,
investment, tuition, etc --- requires domain knowledge.
Furthermore, results grow sparse as more constraints are added.
Figure~\ref{subtable:college-sql} shows results from an SQL
\texttt{SELECT} query with plausible thresholds for this question that yields
only a single match.

This paper shows how to formulate a broad class of probabilistic search queries
on structured data using probabilistic programming and information theory.
The core technical idea combines SQL search operators with a ranking function
called \textit{predictive relevance} that assesses the relevance of database
records to some set of query records, in a context defined by a variable of
interest.
Figures \ref{subtable:college-bql-1} and \ref{subtable:college-bql-2} show two
examples, expanding and then refining the result from %
Figure~\ref{subtable:college-sql} by combining predictive relevance with SQL.
Predictive relevance is the probability that a candidate record is informative
about the answers to a specific class of predictive queries about unknown fields
in the query records.

The paper presents an efficient implementation applying a simple sparse matrix
algorithm to the results of inference in CrossCat
\citep{mansinghka2016}.
The result is a scalable, domain-general search technique for sparse,
multivariate, structured data that combines the strengths of SQL search with
probabilistic approaches to information retrieval.
Users can query by example, using real records in the database if they are
familiar with the domain, or partially-specified hypothetical records if they
are less familiar.
Users can then narrow search results by adding Boolean filters, and by including
multiple records in the query set rather than a single record.
An overview of the technique and its integration into BayesDB
\citep{mansinghka2015} is shown in Figure~\ref{fig:workflow}.

We demonstrate the proposed technique with databases of (i) US colleges, (ii)
public health and macroeconomic indicators, and (iii) cars from the late 1980s.
The paper empirically confirms the scalability of the technique and shows that
human evaluators often prefer results from the proposed technique to results
from a standard baseline.

\input{figures/college}

\FloatBarrier

%% file: figures/college.tex

\begin{figure*}[ht]
\ttfamily\scriptsize

\lstset{
  basicstyle=\ttfamily\scriptsize,
  columns=fullflexible,
  keepspaces=true,
  upquote=true,
  alsoletter={\.,\%},
  morekeywords=[1]{SELECT, FROM, ORDER, BY, RELEVANCE, PROBABILITY,
  EXISTING, ROWS, IN, THE, CONTEXT, OF, TO, AS, WHERE, IS, NOT, DESC, LIMIT,
  SELECT, AND, HYPOTHETICAL, ROW, LIKE, HYPOTHETICAL, WITH, VALUES},
  keywordstyle=[1]\textcolor{DarkGreen},
  showstringspaces=False,
  stringstyle=\ttfamily\color{Sepia},
  morestring=[b]",
}

\begin{subfigure}[t]{.325\linewidth}
\begin{lstlisting}
%bql SELECT
...   "institute",
...   "median_sat_math",
...   "admit_rate",
...   "tuition",
...   "median_student_debt",
...   "instructional_invest",
...   "locale"
... FROM college_scorecard
... WHERE
...   "locale" LIKE '%City%'
...   "tuition" < 50000
...   "median_student_debt" < 10000
...   "instructional_invest" > 50000
... LIMIT 10
\end{lstlisting}
\end{subfigure}%
\begin{subtable}[t]{.675\linewidth}
\subcaption{%
\textbf{Standard SQL}.
Using a SQL \texttt{WHERE} clause to search for a university with a city campus,
low student debt (at most \$10K), high investment in student instruction (at
least \$50K), and a tuition within their budget (at most \$50K).
Due to sparsity in the dataset for the chosen thresholds, the Boolean conditions
in the clause have only a single matching result, shown in the table below.
The user needs to iteratively adjust the thresholds in order to obtain more
results which match the search query.}
\label{subtable:college-sql}
\begin{tabularx}{\linewidth}{Xrrrrrr}
\toprule
institute       & admit & sat & tuition & debt & investment & locale \\
\midrule
Duke University & 11\%  & 745 & 47,243   & 7,500 & 50,756      & Midsize City \\
\bottomrule
\end{tabularx}
\end{subtable}

\begin{subfigure}[t]{.325\linewidth}
\begin{lstlisting}
%bql SELECT
...   "institute",
...   "admit_rate",
...   "median_sat_math",
...   "tuition",
...   "median_student_debt",
...   "instructional_invest",
...   "locale"
... FROM college_scorecard
... ORDER BY
...   RELEVANCE PROBABILITY
...   TO HYPOTHETICAL ROW ((
...    "locale" = 'Midsize City'
...    "tuition" = 50000,
...    "median_student_debt" = 10000,
...    "instructional_invest" = 50000
...   ))
...  IN THE CONTEXT OF
...   "instructional_invest"
... DESC
... LIMIT 10
\end{lstlisting}
\end{subfigure}%
\begin{subtable}[t]{.675\linewidth}
\subcaption{
\textbf{Relevance to hypothetical record}.
If the search query is instead specified as a hypothetical record in a BQL
\texttt{RELEVANCE PROBABILITY} query, then \texttt{ORDER BY} can give the top-10
ranked matches.
The results are all top-tier schools with high teaching investment, a city or
large suburban campus, and low student debt.
However, the user is surprised by the highly stringent admission rates at these
colleges, which are mostly below 10\%.}
\label{subtable:college-bql-1}
\begin{tabularx}{\linewidth}{lrrrrrr}
\toprule
institute             & admit & sat  & tuition & debt  & investment & locale \\
\midrule
Duke University       & 11\%  & 745  & 47,243   & 7,500  & 50,756      & Midsize City \\
Princeton University  & 8\%   & 755  & 41,820   & 7,500  & 52,224      & Large Suburb \\
Harvard University    & 6\%   & 755  & 43,938   & 6,500  & 49,500      & Midsize City \\
Univ of Chicago       & 8\%   & 758  & 49,380   & 12,500 & 83,779      & Large City \\
Mass Inst Technology  & 8\%   & 770  & 45,016   & 14,990 & 62,770      & Midsize City \\
Calif Inst Technology & 8\%   & 785  & 43,362   & 11,812 & 92,590      & Midsize City \\
Stanford University   & 5\%   & 745  & 45,195   & 12,782 & 93,146      & Large Suburb \\
Yale University       & 6\%   & 750  & 45,800   & 13,774 & 107,982     & Midsize City\\
Columbia University   & 7\%   & 745  & 51,008   & 23,000 & 80,944      & Large City \\
University of Penn.   & 10\%  & 735  & 47,668   & 21,500 & 49,018      & Large City \\
\bottomrule
\end{tabularx}

\end{subtable}

\begin{subfigure}[t]{.325\linewidth}
\begin{lstlisting}
%bql SELECT
...   "institute",
...   "admit_rate",
...   "median_sat_math",
...   "tuition",
...   "median_student_debt",
...   "instructional_invest",
...   "locale"
... FROM college_scorecard
... WHERE
...   "admit_rate" > 0.10
...   AND "locale" LIKE '%City%'
... ORDER BY
...   RELEVANCE PROBABILITY
...   TO EXISTING ROWS IN (
...    'Duke University',
...    'Harvard University',
...    'Mass Inst Technology',
...    'Yale University',
...   )
...  IN THE CONTEXT OF
...   "instructional_invest"
... DESC
... LIMIT 10
\end{lstlisting}
\end{subfigure}%
\begin{subtable}[t]{.675\linewidth}
\subcaption{%
\textbf{Relevance to observed records combined with SQL}.
Combining BQL and SQL to search for colleges which are most relevant to the
schools from \subref{subtable:college-bql-1} in the context of
``instructional investment'', but that must have (i) less stringent admissions
(at least 10\%) and (ii) city campuses only.
The quantitative search metrics of interest for the colleges in the result set
are all significantly better than the national average, but they are mostly
below the more selective schools in \subref{subtable:college-bql-1}.}
\label{subtable:college-bql-2}
\begin{tabularx}{\linewidth}{Xrrrrrr}
\toprule
institute             & admit & sat    & tuition & debt    & investment & locale \\
\midrule
Duke University     & 11\% & 745 & 47,243 & 7,500  & 50,756 & Midsize City \\
Georgetown Univ     & 17\% & 710 & 46,744 & 17,000 & 31,102 & Midsize City \\
Johns Hopkins Univ  & 16\% & 730 & 47,060 & 16,250 & 77,339 & Midsize City  \\
Vanderbilt Univ     & 13\% & 760 & 43,838 & 13,000 & 79,372 & Large City \\
University of Penn. & 10\% & 735 & 47,668 & 21,500 & 49,018 & Large City \\
Carnegie Mellon     & 24\% & 750 & 49,022 & 25,250 & 31,807 & Midsize City \\
Rice University     & 15\% & 750 & 40,566 & 9,642  & 40,056 & Midsize City \\
Univ Southern Calif & 18\% & 710 & 48,280 & 21,500 & 43,170 & Midsize City \\
Cooper Union        & 15\% & 710 & 41,400 & 18,250 & 21,635 & Large City \\
New York University & 35\% & 685 & 46,170 & 23,300 & 30,237 & Large City \\
\bottomrule
\end{tabularx}
\includegraphics[width=\linewidth]{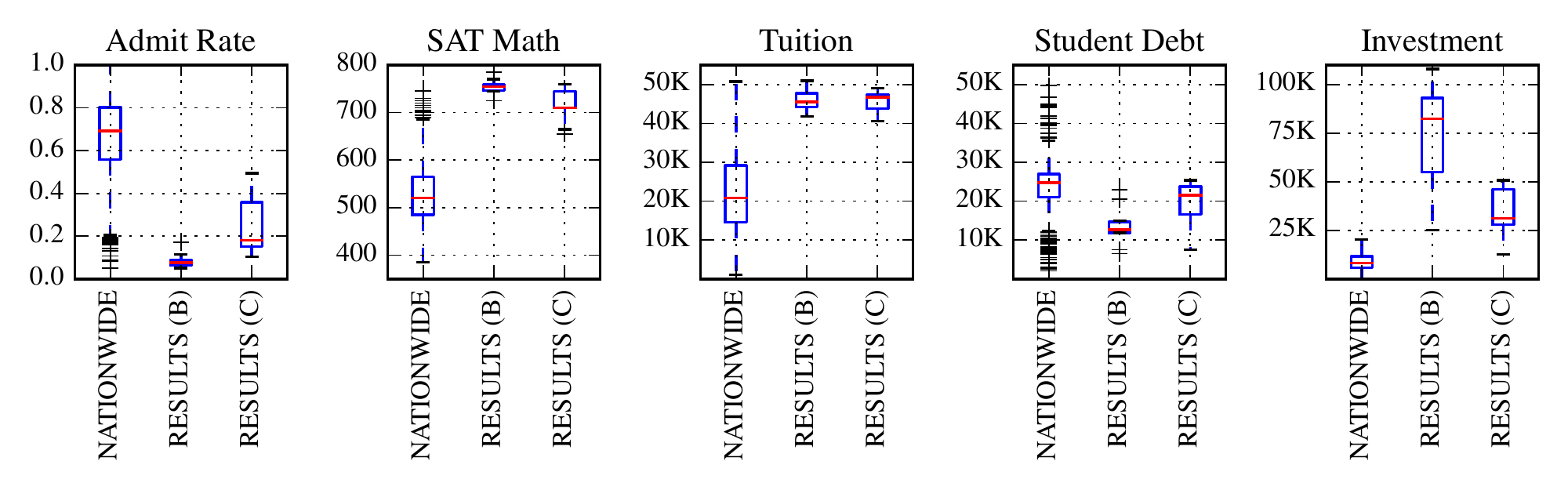}
\end{subtable}

\caption{Combining predictive relevance probability in the Bayesian Query
Language (BQL) with standard techniques in SQL to search the US College
Scorecard dataset. The full data contains over 7000 colleges and 1700 variables,
and is available for download at
\mbox{\url{collegescorecard.ed.gov/data}}.}
\label{fig:college}
\end{figure*}

%% file: bayesian-relevance.tex

\section{Establishing an information theoretic definition of context-specific
predictive relevance}
\label{sec:establishing-mi}

\input{figures/graphical}
\input{figures/workflow}

In this section, we outline the basic set-up and notations for the database
search problem, and establish a formal definition of the probability of
``predictive relevance'' between records in the database.

\subsection{Finding predictively relevant records}
\label{subsec:finding-records}

Suppose we are given a sparse dataset $\D\,{=}\,\set{\x_1, \x_2, \dots,
\x_N}$ containing $N$ records, where each
$\x_r\,{=}\,(x_{[r,1]},\dots,x_{[r,p]})$ is an instantiation of a
$p$-dimensional random vector, possibly with missing values.
For notational convenience, we refer to arbitrary collections of observations
using sets as indices, so that %
$\x_{[R,C]}\,{\equiv}\,\set{x_{[r,c]} : r\in{R}, c\in{R}}$. Bold-face symbols
denote multivariate entities, and variables are capitalized as $X_{[r,c]}$ when
they are unobserved (i.e. random).

Let $\mtQ\,{\subset}\,[N]$ index a small collection of ``query records''
$\x_\mtQ\,{=}\,\set{\x_q : q \in \mtQ}$.
Our objective is to rank each item $\x_i\,{\in}\,\D$ by how relevant it is for
formulating predictions about values of $\x_\mtQ$, ``in the context'' of a
particular dimension $c$.
We formally define the context of $c$ as a subset of dimensions
$\mtV\,{\subseteq }\,[p]$ such that for an arbitrary record $r^*$ and each
$v\,{\in}\,\mtV$, the random variable $X_{[r^*,v]}$ is statistically dependent
with $X_{[r^*,c]}$.\footnote{%
A general definition for statistical dependence is having non-zero mutual
information with the context variable. However, the method for detecting
dependence to find variables in the context can be arbitrary e.g., using linear
statistics such as Pearson-R, directly estimating mutual information, or
others.}

In other words, we are searching for records $i$ where knowledge of
$\x_{[i,\mtV]}$ is useful for predicting $\x_{[\mtQ,\mtV]}$, had we not known
the values of these observations.

\subsection{Defining context-specific predictive relevance using mutual
information}
\label{subsec:defining-relevance}

We now formalize the intuition from the previous section more precisely.
Let $\mtR_c(\mtQ, r)$ denote the probability that $r$ is predictively relevant
to $\mtQ$, in the context of $c$.
Furthermore, let $c^*$ denote the index of a new dimension in the length-$p$
random vectors, which is statistically dependent on dimension $c$ (i.e. is in
its context) but is not one of the $p$ existing variables in the database.
Since $c^*$ indexes a novel variable, its value for each row $r$ is itself
a random variable, which we denote $X_{[r,c^*]}$.
We now define the probability that $r$ is predictively relevant to $\mtQ$ in the
context of $c$ as the posterior probability that the mutual information of
$X_{[r,c^*]}$ and each query record $X_{[q,c^*]}$ is non-zero:
\begin{flalign}
&\mtR_c(\mtQ,r) = \label{eq:def-relevance} &&\\
&\quad\quad \mathbb{P}\left[
  \bigcap\limits_{q\in\mtQ} \left(
    \mtI(X_{[q,c^*]} : X_{[r,c^*]}) > 0
  \right) \mathrel{\bigg|} \lambda_{c^*}, \alpha, \D
  \right]. \notag
\end{flalign}
The symbol $\lambda_{c^*}$ refers to an arbitrary set of hyperparameters which
govern the distribution of dimension $c^*$, and $\alpha$ is a context-specific
hyperparameter which controls the prior on structural dependencies between the
random variables $\set{X_{[r,c^*]} : r \in [N]}$.
Moreover, the mutual information $\mtI$, a well-established
measure for the strength of predictive relationships between
random variables \citep{cover2012}, is defined in the usual way,
\begin{flalign}
&\mtI(X_{[q,c^*]} : X_{[r,c^*]}\mid\lambda_{c^*},\alpha,\D)= && \label{eq:def-mi}\\
&\quad \mathbb{E}\left[
 \log\left(
    \frac{
      p(X_{[q,c^*]}, X_{[r,c^*]}|\lambda_{c^*},\alpha,\D)}{
      p(X_{[q,c^*]}|\lambda_{c^*},\alpha,\D)
      p(X_{[r,c^*]}|\lambda_{c^*},\alpha,\D)
    }
 \right)
\right]. \notag
\end{flalign}
Figure~\ref{fig:hypothesis} illustrates the predictive relevance probability in
terms of a hypothesis test on two competing graphical models, where the mutual
information is non-zero in panel~\subref{subfig:hypothesis-same} indicating
predictive relevance; and zero in panel~\subref{subfig:hypothesis-diff},
indicating predictive irrelevance.

\subsection{Related Work}
\label{subsec:related-work}

Our formulation of predictive relevance in terms of mutual information between
new variables $X_{[r,c^*]}$ is related to the idea of ``property induction''
from the cognitive science literature
\citep{rips1975,osherson1990,shafto2008}, where subjects are asked to predict
whether an entity has a property, given that some other entity has that
property; e.g. how likely are cats to have some new disease, given that mice are
known to have the disease?


It is also informative to consider the relationship between the predictive
relevance $\mtR_c(\mtQ,r)$ in Eq~\eqref{eq:def-relevance} and the Bayesian Sets
ranking function from the statistical modeling literature
\citep{ghahramani2005}:
\begin{align}
\textrm{score}_{\textrm{Bayes-Sets}}(\mtQ,r) =
\frac{p(\x_r{\mid}\x_\mtQ)}{p(\x_r)}.
\label{eq:bayesian-sets}
\end{align}
Bayes Sets defines a Bayes Factor, or ratio of marginal likelihoods, which is
used for hypothesis testing without assuming a structure prior.
On the other hand, predictive relevance defines a posterior probability, whose
value is between 0 and 1, and therefore requires a prior over dependence
structure between records (our approach outlined in Section~\ref{sec:crosscat}
is based on nonparametric Bayes).
While Bayes Sets draws inferences using only the query and candidate rows
without considering the rest of the data, predictive relevance probabilities are
necessarily conditioned on $\D$ as in Eq~\eqref{eq:def-relevance}.
Finally Bayes Sets considers the entire data vectors for scoring, whereas
predictive relevance considers only dimensions which are in the context of a
variable $c$, making it possible for two records to be predictively relevant in
some context but probably predictively irrelevant in another.

%% file: figures/graphical.tex

\begin{figure*}[ht]
\centering
\begin{subfigure}[b]{.35\linewidth}
\includestandalone[width=\textwidth]{figures/relevance-mi-hypotheses-same}%
\caption{Same generative process for $\x_\mtQ$ and $x_r$.}
\label{subfig:hypothesis-same}
\end{subfigure}\qquad%
\begin{subfigure}[b]{.4\linewidth}
\includestandalone[width=\textwidth]{figures/relevance-mi-hypotheses-diff}
\caption{Different generative processes for $\x_\mtQ$ and $x_r$.}
\label{subfig:hypothesis-diff}
\end{subfigure}%
\caption{%
The predictive relevance of a collection of query records $\mtQ$ to a candidate
record $r$, in the context of variable $c$, computes the probability that
$\x_{[\mtQ,c]}$ and $x_{[r,c]}$ are drawn from \subref{subfig:hypothesis-same}
the same generative process, versus \subref{subfig:hypothesis-diff} different
generative processes.
The latent variables $z_0$ and $z_1$ are indicators for the generative process
of the records; and $\theta^c_0$ (resp. $\theta^c_1)$ are distributional
parameters of data under model $z_0$ (resp. $z_1$) for variable $c$.
Hyperparameter $\alpha$ dictates the prior on $z$, and $\lambda$
dictates the prior on distributional parameters $\theta$.
The symbol $c^*$ denotes a new dimension which is statistically dependent on
$c$, and for which no values are observed for either $\mtQ$ or $r.$
Conditioned on hyperparameters, knowing $X_{[r,c^*]}$ in
\subref{subfig:hypothesis-same} carries information about the unknown values
$\X_{[\mtQ,c^*]}$, whereas in \subref{subfig:hypothesis-diff} it does not.%
}
\label{fig:hypothesis}
\end{figure*}

%% file: figures/relevance-mi-hypotheses-same.tex
\begin{tikzpicture}[transform shape]


\node[
  const,
] (alpha_v) {
  $\alpha$
};


\node[
  latent,
  below = of alpha_v,
] (z_q) {
  $z_0$
};

\node[
  latent,
  below left = 1 and 1.5 of z_q,
] (theta_qc) {
  $\theta_{0}^c$
};

\node[
  latent,
  below right = 1 and 1.5 of z_q,
] (theta_qcs) {
  $\theta_{0}^{c^*}$
};

\node[
  obs,
  minimum size = 10pt,
  below = of theta_qc,
] (obsq-1) {
};

\node[
  obs,
  minimum size = 10pt,
  left = 0.35 of obsq-1,
] (obsq-2) {
};

\node[
  rectangle,
  minimum size = 10pt,
  right = 0.0 of obsq-1,
] (obsq-dots) {
  $\dots$
};

\node[
  obs,
  minimum size = 10pt,
  right = 0.70 of obsq-1,
] (obsq-3) {
};

\node[
  obs,
  minimum size = 10pt,
  right = 0.35 of obsq-3,
] (obsr) {
};

\node[
  latent,
  minimum size = 10pt,
  below = of theta_qcs,
] (hypq-1) {
};

\node[
  latent,
  minimum size = 10pt,
  left = 0.35 of hypq-1,
] (hypq-2) {
};

\node[
  rectangle,
  minimum size = 10pt,
  right = 0.0 of hypq-1,
] (hypq-dots) {
  $\dots$
};

\node[
  latent,
  minimum size = 10pt,
  right = 0.70 of hypq-1,
] (hypq-3) {
};

\node[
  latent,
  minimum size = 10pt,
  right = 0.35 of hypq-3,
] (hypr) {
};

\draw [
    thick,
    decoration={
        brace,
        mirror,
        raise=0.25cm
    },
    decorate
] (obsq-2.west) -- (obsq-3.east);

\node[
  rectangle,
  yshift = -0.65cm,
] (x_qc) at ($(obsq-2.west)!0.5!(obsq-3.east)$) {
  $\x_{[Q,c]}$
};

\draw [
    thick,
    decoration={
        brace,
        mirror,
        raise=0.25cm
    },
    decorate
] (hypq-2.west) -- (hypq-3.east);

\node[
  rectangle,
  yshift = -0.65cm,
] (x_qc) at ($(hypq-2.west)!0.5!(hypq-3.east)$) {
  $\X_{[Q,c^*]}$
};

\node[
  rectangle,
] (x_rc) at (x_qc -| obsr) {
  $x_{[r,c]}$
};

\node[
  rectangle,
] (x_rc) at (x_qc -| hypr) {
  $x_{[r,c^*]}$
};


\edge[-latex] {alpha_v} {z_q};

\edge[-latex] {z_q} {theta_qc};
\edge[-latex] {z_q} {theta_qcs};

\edge[-latex] {theta_qc} {obsq-1};
\edge[-latex] {theta_qc} {obsq-2};
\edge[-latex] {theta_qc} {obsq-3};
\edge[-latex] {theta_qc} {obsr};

\edge[-latex] {theta_qcs} {hypq-1};
\edge[-latex] {theta_qcs} {hypq-2};
\edge[-latex] {theta_qcs} {hypq-3};
\edge[-latex] {theta_qcs} {hypr};


\node[
  const,
  right = of alpha_v,
] (lambda_cs) {
  $\lambda_{c^*}$
};

\node[
  const,
  left = of alpha_v,
] (lambda_c) {
  $\lambda_{c}$
};

\draw[-latex, dotted] (lambda_c.south) -- (theta_qc);
\draw[-latex, dotted] (lambda_cs.south) -- (theta_qcs);

\end{tikzpicture}

%% file: figures/relevance-mi-hypotheses-diff.tex
\begin{tikzpicture}[transform shape]


\node[
  const,
] (alpha_v) {
  $\alpha$
};


\node[
  latent,
  below left = 1 and 2 of alpha_v,
] (z_q) {
  $z_0$
};

\node[
  latent,
  below left = of z_q,
] (theta_qc) {
  $\theta_{0}^c$
};

\node[
  latent,
  below right = of z_q,
] (theta_qcs) {
  $\theta_{0}^{c^*}$
};

\node[
  obs,
  minimum size = 10pt,
  below = of theta_qc,
] (obsq-1) {
};

\node[
  obs,
  minimum size = 10pt,
  left = 0.35 of obsq-1,
] (obsq-2) {
};

\node[
  rectangle,
  minimum size = 10pt,
  right = 0.0 of obsq-1,
] (obsq-dots) {
  $\dots$
};

\node[
  obs,
  minimum size = 10pt,
  right = 0.70 of obsq-1,
] (obsq-3) {
};

\node[
  latent,
  minimum size = 10pt,
  below = of theta_qcs,
] (hypq-1) {
};

\node[
  latent,
  minimum size = 10pt,
  left = 0.35 of hypq-1,
] (hypq-2) {
};

\node[
  rectangle,
  minimum size = 10pt,
  right = 0.0 of hypq-1,
] (hypq-dots) {
  $\dots$
};

\node[
  latent,
  minimum size = 10pt,
  right = 0.70 of hypq-1,
] (hypq-3) {
};

\draw [
    thick,
    decoration={
        brace,
        mirror,
        raise=0.25cm
    },
    decorate
] (obsq-2.west) -- (obsq-3.east);

\node[
  rectangle,
  yshift = -0.65cm,
] (x_qc) at ($(obsq-2.west)!0.5!(obsq-3.east)$) {
  $\x_{[\mtQ,c]}$
};

\draw [
    thick,
    decoration={
        brace,
        mirror,
        raise=0.25cm
    },
    decorate
] (hypq-2.west) -- (hypq-3.east);

\node[
  rectangle,
  yshift = -0.65cm,
] (x_qc) at ($(hypq-2.west)!0.5!(hypq-3.east)$) {
  $\X_{[\mtQ,c^*]}$
};


\node[
  latent,
  below right = 1 and 2 of alpha_v,
] (z_r) {
  $z_1$
};

\node[
  latent,
  below left = of z_r,
] (theta_rc) {
  $\theta_{1}^c$
};

\node[
  latent,
  below right = of z_r,
] (theta_rcs) {
  $\theta_{1}^{c^*}$
};

\node[
  obs,
  minimum size = 10pt,
  below = of theta_rc,
] (obsr-1) {
};

\node[
  rectangle,
] (x_rc) at (x_qc -| obsr-1) {
  $x_{[r,c]}$
};

\node[
  latent,
  minimum size = 10pt,
  below = of theta_rcs,
] (hypr-1) {
};

\node[
  rectangle,
] (x_rc) at (x_qc -| hypr-1) {
  $X_{[r,c^*]}$
};


\node[
  const,
  left = of alpha_v,
] (lambda_cs) {
  $\lambda_{c^*}$
};

\node[
  const,
  left = of lambda_cs,
] (lambda_c) {
  $\lambda_{c}$
};

\draw[-latex, dotted] (lambda_c.south) -- (theta_rc);
\draw[-latex, dotted] (lambda_c.south) -- (theta_qc);
\draw[-latex, dotted] (lambda_cs.south) -- (theta_rcs);
\draw[-latex, dotted] (lambda_cs.south) -- (theta_qcs);


\edge[-latex] {alpha_v} {z_q};
\edge[-latex] {alpha_v} {z_r};

\edge[-latex] {z_q} {theta_qc};
\edge[-latex] {z_q} {theta_qcs};

\edge[-latex] {z_r} {theta_rc};
\edge[-latex] {z_r} {theta_rcs};

\edge[-latex] {theta_qc} {obsq-1};
\edge[-latex] {theta_qc} {obsq-2};
\edge[-latex] {theta_qc} {obsq-3};

\edge[-latex] {theta_qcs} {hypq-1};
\edge[-latex] {theta_qcs} {hypq-2};
\edge[-latex] {theta_qcs} {hypq-3};

\edge[-latex] {theta_rc} {obsr-1};
\edge[-latex] {theta_rcs} {hypr-1};

\end{tikzpicture}

%% file: figures/workflow.tex


\begin{figure*}[ht]
\centering
\begin{tikzpicture}

\renewcommand{\ttdefault}{cmtt}


\node[
    rectangle,
    draw = none,
    align = center
] (data-table-header) {\footnotesize
    \textbf{Sparse Tabular Database}
};

\node[
  rectangle,
  draw = none,
  align = center,
  minimum width = 6cm,
  below = 0cm of data-table-header
] (data-table){
  \scriptsize
  \begin{tabular}{|l|l|l|l|l|}
  \hline
  \textbf{country}
    & \textbf{oil}
    & \textbf{hdi}
    & \textbf{snow}
    & \textbf{government}
    \\ \hline
  Australia       &  19         &            &            & parliamentary   \\
  Lebanon         &             & 145        & 1.3        & semi-presidential      \\
  Swaziland       &  17         & 110        &            & monarchy    \\
  USA             &  31         & 197        & 2.9        & presidential  \\
  China           &  21         &            & 3.4        & politburo      \\
  Greece          &  03         & 180        &            & parliamentary    \\
  Peru            &             & 147        & 1.1        & presidential  \\
  \dots           &  \dots      & \dots      & \dots      & \dots     \\
  \hline
  \end{tabular}
};


\node[
    rectangle,
    draw = none,
    align = center,
    right = 1.2 cm of data-table-header
] (bayesdb-modeling) {\footnotesize
    \textbf{BayesDB Modeling}
};

\draw [-stealth,thick] (data-table-header) -- (bayesdb-modeling);


\node[
    rectangle,
    draw = none,
    align = center,
    right = 1.2cm of bayesdb-modeling
] (crosscat-header) {\footnotesize
    \textbf{Posterior CrossCat Structures}
};

\draw [-stealth,thick] (bayesdb-modeling) -- (crosscat-header);

\node[
  rectangle,
  draw = none,
  align = center,
  right = 4.5cm of data-table.south,
  anchor = south,
] (crosscat-structure-1){
  \tiny

  \begin{tabular}{|>{\columncolor{gray!40}}c|}
  \hline
  \multicolumn{1}{|c|}{\textbf{O}}\\
  \hline
  \\
  \\
  \arrayrulecolor{red}\hline
  \\
  \\
  \\
  \arrayrulecolor{black}\hline
  \end{tabular}

  \begin{tabular}{|>{\columncolor{gray!40}}c|>{\columncolor{gray!40}}c|}
  \hline
  \multicolumn{1}{|c|}{\textbf{H}} & \multicolumn{1}{c|}{\textbf{S}}\\
  \hline
  & \\
  \arrayrulecolor{red}\hline
  & \\
  & \\
  \arrayrulecolor{red}\hline
  & \\
  & \\
  \arrayrulecolor{black}\hline
  \end{tabular}

  \begin{tabular}{|>{\columncolor{gray!40}}c|}
  \hline
  \multicolumn{1}{|c|}{\textbf{G}}\\
  \hline
  \\
  \\
  \\
  \\
  \arrayrulecolor{red}\hline
  \\
  \arrayrulecolor{black}\hline
  \end{tabular}
};

\node[
  rectangle,
  draw = none,
  align = center,
  right = .4cm of crosscat-structure-1
] (crosscat-structure-2){
  \tiny

  \begin{tabular}{|>{\columncolor{gray!40}}c|>{\columncolor{gray!40}}c|}
  \hline
  \multicolumn{1}{|c|}{\textbf{O}} & \multicolumn{1}{c|}{\textbf{H}}\\
  \hline
  & \\
  & \\
  & \\
  \arrayrulecolor{red}\hline
  & \\
  & \\
  \arrayrulecolor{black}\hline
  \end{tabular}

  \begin{tabular}{|>{\columncolor{gray!40}}c|}
  \hline
  \multicolumn{1}{|c|}{\textbf{S}}\\
  \hline
  \\
  \arrayrulecolor{red}\hline
  \\
  \\
  \\
  \\
  \arrayrulecolor{black}\hline
  \end{tabular}

  \begin{tabular}{|>{\columncolor{gray!40}}c|}
  \hline
  \multicolumn{1}{|c|}{\textbf{G}}\\
  \hline
  \\
  \\
  \\
  \\
  \arrayrulecolor{red}\hline
  \\
  \arrayrulecolor{black}\hline
  \end{tabular}
};

\node[
  rectangle,
  draw = none,
  align = center,
  right = .4cm of crosscat-structure-2
] (crosscat-structure-3){
  \tiny
  \begin{tabular}{%
    |>{\columncolor{gray!40}}c%
    |>{\columncolor{gray!40}}c%
    |>{\columncolor{gray!40}}c%
    |>{\columncolor{gray!40}}c|}
  \hline
  \multicolumn{1}{|c|}{\textbf{O}}
    & \multicolumn{1}{c|}{\textbf{H}}
    & \multicolumn{1}{c|}{\textbf{S}}
    & \multicolumn{1}{c|}{\textbf{G}} \\
  \hline
  & & & \\
  \arrayrulecolor{red}\hline
  & & & \\
  & & & \\
  \arrayrulecolor{red}\hline
  & & & \\
  & & & \\
  \arrayrulecolor{black}\hline
  \end{tabular}%
};


\node[
    rectangle,
    align = center,
    above = 0 of crosscat-structure-1.90,
] (crosscat-structure-1-header) {\footnotesize
    Model $\hat{\bphi}^1$
};

\node[
    rectangle,
    align = center,
    above = 0 of crosscat-structure-2.90,
] (crosscat-structure-2-header) {\footnotesize
    Model $\hat{\bphi}^2$
};

\node[
    rectangle,
    align = center,
    above = 0 of crosscat-structure-3.90,
] (crosscat-structure-3-header) {\footnotesize
    Model $\hat{\bphi}^3$
};

\draw[-stealth,thick]
    (crosscat-header.south) -- (crosscat-structure-1-header.north);

\draw[-stealth,thick]
    (crosscat-header.south) -- (crosscat-structure-2-header.north);

\draw[-stealth,thick]
    (crosscat-header.south) -- (crosscat-structure-3-header.north);




\node[
    rectangle,
    draw = none,
    align = center,
    below = 0.25cm of data-table.south
] (bql-query-header) {\footnotesize
    \textbf{BQL Predictive Relevance Query}
};

\node[
    rectangle,
    draw = black,
    align = left,
    minimum width = 6cm,
    below = 0cm of bql-query-header.south
] (bql-query) {
    \lstset{
      basicstyle=\ttfamily\scriptsize,
      columns=fullflexible,
      keepspaces=true,
      upquote=true,
      alsoletter={\.,\%},
      morekeywords=[1]{SELECT, FROM, ORDER, BY, PREDICTIVE, RELEVANCE,
      PROBABILITY, EXISTING, ROWS, IN, THE, CONTEXT, OF, TO, AS, WHERE, IS, NOT,
      NEW, ROW, WITH, VALUES, RECORD, HYPOTHETICAL},
      keywordstyle=[1]\textcolor{DarkGreen},
      showstringspaces=False,
      stringstyle=\ttfamily\color{Sepia},
      morestring=[b]",
    }
  \begin{lstlisting}
%bql SELECT "country", "oil", "hdi"
...  FROM population
...  WHERE "government" IS NOT 'monarchy'
...  ORDER BY
...    RELEVANCE PROBABILITY
...    TO HYPOTHETICAL ROW WITH VALUES
...      (("oil"=27, "snow"=0.2, "hdi"=180))
...    IN THE CONTEXT OF "hdi"
  \end{lstlisting}
};


\node[
    rectangle,
    draw = none,
    align = center,
    below = 0.25cm of bql-query.south
] (bql-results-header) {\footnotesize
    \textbf{Query Results}
};

\node[
    rectangle,
    align = left,
    below = 0cm of bql-results-header.south
] (bql-results) {
  \ttfamily\scriptsize
  \hspace{.65cm}\begin{tabular}{|l|c|c|}
  \hline
  country & oil & hdi \\
  \hline
  USA             &  31         & 197     \\
  Australia       &  19         &         \\
  Greece          &  03         & 180     \\
  Peru            &  17         & 147     \\
  China           &  21         &         \\
  Lebanon         &             & 145     \\
  \dots           &  \dots      & \dots   \\
  \hline
  \end{tabular}
};


\node[
    rectangle,
    draw = black,
    fill = gray!40!white,
    align = center,
    minimum width = 7cm,
] (bayesdb-relevance-estimator)
  at (bql-query -| crosscat-structure-2) {
    \footnotesize\textsc{CrossCat-Incorporate-Record}
      (Algorithm~\ref{alg:crosscat-incorporate})\\
    \footnotesize\textsc{CrossCat-Predictive-Relevance}
      (Algorithm~\ref{alg:crosscat-relevance})
};

\node[
    rectangle,
    align = center,
    minimum width = 4cm,
    above = 0 of bayesdb-relevance-estimator.north,
] (bayesdb-relevance-estimator-header) {\footnotesize
    \textbf{BayesDB Query Engine}
};

\draw[-stealth,thick]
    (bql-query) -- (bayesdb-relevance-estimator);

\draw[-stealth,]
    (crosscat-structure-1.south) -- (bayesdb-relevance-estimator-header);

\draw[-stealth]
    (crosscat-structure-2.south) -- (bayesdb-relevance-estimator-header);

\draw[-stealth]
    (crosscat-structure-3.south) -- (bayesdb-relevance-estimator-header);


\node[
  rectangle,
  anchor=south,
] (crosscat-scores)
  at (bql-results.south -| bayesdb-relevance-estimator) {
  \scriptsize
  \begin{tabular}{|l|l|l|l|l|}
  \hline
  \multirow{2}{*}{\textbf{Country}}
    & \multicolumn{4}{c|}{\textbf{Relevance Prob}}
  \\
    & \multicolumn{1}{c}{$\hat\bphi^1$}
    & \multicolumn{1}{c}{$\hat\bphi^2$}
    & \multicolumn{1}{c}{$\hat\bphi^3$}
    & \multicolumn{1}{c|}{avg}
  \\
  \hline
  China       & 0   & 1 & 0 & 0.33 \\
  USA         & 1   & 1 & 1 & 1.00 \\
  Lebanon     & 0   & 0 & 0 & 0.00 \\
  Greece      & 1   & 0 & 1 & 0.66 \\
  Australia   & 1   & 0 & 1 & 0.66 \\
  Peru        & 1   & 0 & 0 & 0.33 \\
  \dots       & \dots & \dots & \dots & \\
  \hline
  \end{tabular}
};


\node[
    rectangle,
    draw = none,
    align = center,
    left = .75cm of crosscat-scores
] (bayesdb-averaging) {
  \footnotesize\textbf{SQL}\\
  \footnotesize\textbf{Sorting}
};

\coordinate[
  left = 1.25 of bayesdb-averaging.west
] (bayesdb-averaging-anchor);

\draw[-stealth, thick]
  (bayesdb-relevance-estimator.south) -- (crosscat-scores.north);

\draw[-stealth, thick] (crosscat-scores) -- (bayesdb-averaging);
\draw[-stealth, thick] (bayesdb-averaging) -- (bayesdb-averaging-anchor);

\end{tikzpicture}

\caption{%
BayesDB workflow for computing context-specific predictive relevance between
database records. Modeling and inference in BayesDB produces an ensemble of
posterior CrossCat model structures. Each structure specifies (i) a column
partition for the factorization of the joint distribution of all variables in
the database, using a Chinese restaraunt process; and (ii) a separate row
partition within each block of variables, using a Dirichlet process mixture. The
column partition clusters variables into different ``contexts'', where all
variables in a context are probably dependent on one another. With each context,
the row partition clusters records which are probably informative of one
another. \mbox {End-user} queries for predictive relevance are expressed in
Bayesian Query Langauge. The BQL interpreter aggregates relevance probabilities
across the ensemble, and can use them as a ranking function in a probabilistic
\texttt{ORDER BY} query.}
\label{fig:workflow}

\end{figure*}
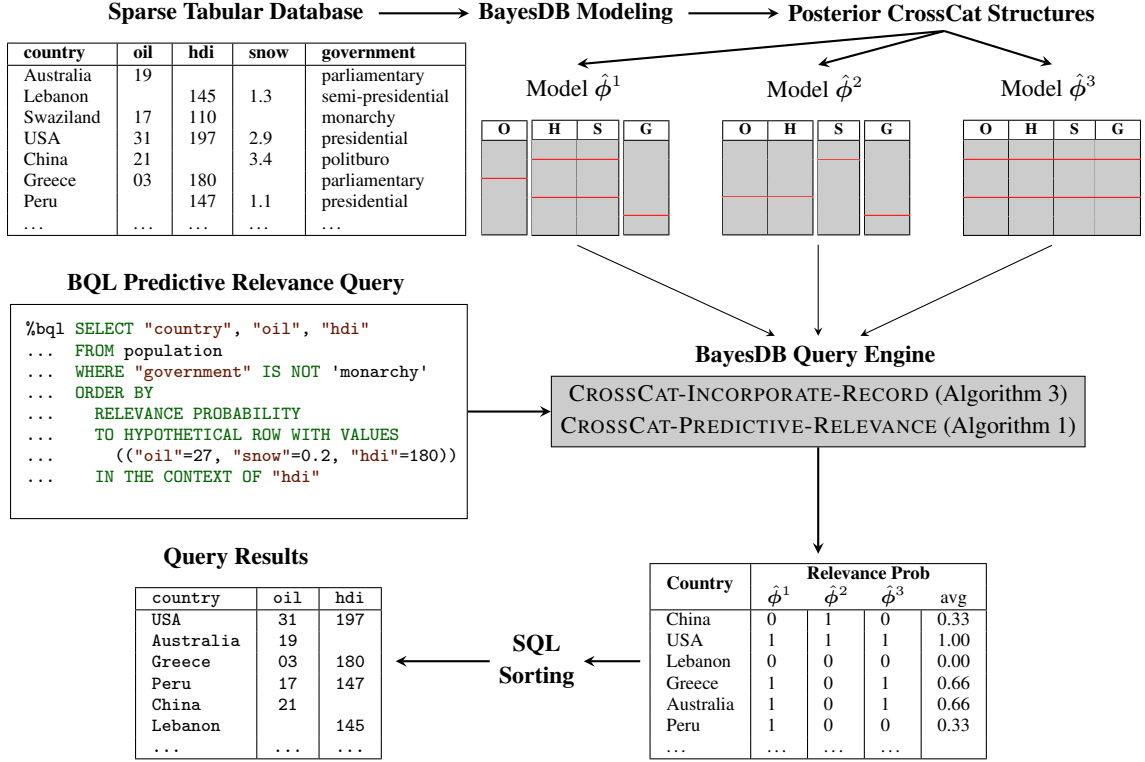

%% file: crosscat-relevance.tex

\section{Computing the probability of predictive relevance using nonparametric
Bayes}
\label{sec:crosscat}

This section describes the cross-categorization prior (CrossCat,
\citet{mansinghka2016}) and outlines algorithms which use CrossCat to
efficiently estimate predictive relevance probabilities %
Eq~\eqref{eq:def-relevance} for sparse, high-dimensional, and %
heterogenously-typed data tables.

CrossCat is a nonparametric Bayesian model which learns the full joint
distribution of $p$ variables using structure learning and %
divide-and-conquer.
The generative model begins by partitioning the set of $p$ variables into blocks
using a Chinese restaurant process.
This step is CrossCat's ``outer'' clustering, since it partitions the columns of
a data table where variables correspond to columns, and records correspond to
rows.
Let $\pi$ denote the partition of $[p]$ whose $k$-th block is
$\mtV^k\,{\subseteq}\,[p]$: for $j\,{\ne}\,{k}$, all variables in $\mtV^k$ are
mutually (marginally and conditionally) independent of all variables in
$\mtV^j$.
Within block $k$, the variables $\x_{[r,\mtV^k]}$ follow a Dirichlet process
mixture model \citep{escobar1995}, where we focus on the case the joint
distribution factorizes given the latent cluster assignment $z^k_r$.
This step is an ``inner'' clustering in CrossCat, since it specifies a cluster
assignment for each row in block $k$.
CrossCat's combinatorial structure requires detailed notation to track the
latent variables and dependencies between them.
The generative process for an exchangeable sequence $(\X_1,\dots,\X_N)$ of $N$
random vectors is summarized below.
\begin{table}[H]

\footnotesize

\caption{Symbols used to describe CrossCat prior}
\label{tab:crosscat-symbols}
\begin{tabularx}{\linewidth}{|l|X|}
\hline
\textbf{Symbol} & \textbf{Description} \\ \hline
$\alpha_0$ & Concentration hyperparameter of column CRP \\
$\alpha_1$ & Concentration hyperparameter of row CRP \\
$v_c$ & Index of variable $c$ in column partition\\
$\mtV^k$ & List of variables in block $k$ of column partition \\
$z^k_r$ & Cluster index of $r$ in row partition of block $k$ \\
$\mtC^k_y$ & List of rows in cluster $y$ of block $k$ \\
$M_c$ & Joint distribution of data for variable $c$ \\
$\lambda_c$ & Hyperparameters of $M_c$ \\
$X_{[r,c]}$ & $r$-{th} observation of variable $c$ \\
$\textsc{\small Set}(l)$ & Unique items in list $l$\\
\hline
\end{tabularx}

\medskip

\begin{mdframed}
\underline{\textsc{CrossCat Prior}}

\medskip

\begin{tabular*}{\linewidth}{ll}
\multicolumn{2}{@{}l}
  {1. {Sample column partition into blocks.}}
  \\[3pt]
$\bv = (v_1,\dots,v_p) \sim \textsc{\small Crp}(\cdot|\alpha_0)$
  & \\
$\mtV^k \gets \set{c\,{\in}\,[p]: v_c\,{=}\,k}$
  & foreach $k\,{\in}\,\textsc{\small Set}(\bv)$
  \\[5pt]
\multicolumn{2}{@{}l}
  {2. {Sample row partitions within each block.}}
  \\[3pt]
$\z^k\,{=}\,(z^k_1,\dots,z^k_N) \sim \textsc{\small Crp}(\cdot|\alpha_1)$
  & foreach $k\,{\in}\,\textsc{\small Set}(\bv)$
  \\
$\mtC^k_y \gets \set{r\,{\in}\,[N]: z^k_r\,{=}\,y}$
  & foreach $k\,{\in}\,\textsc{\small Set}(\bv)$
  \\
$\phantom{!}$
  & \;foreach $y\,{\in}\,\textsc{\small Set}(\z^k)$
  \\[5pt]
\multicolumn{2}{@{}l}
  {3. {Sample data jointly within row cluster.}}
  \\[3pt]
$\set{X_{[r,c]} : r \in \mtC^k_y} \sim M_c(\cdot|\lambda_c)$
  & foreach $k\,{\in}\,\textsc{\small Set}(\bv)$
  \\
$\phantom{!}$
  & \;foreach $y\,{\in}\,\textsc{\small Set}(\z^k)$
  \\
$\phantom{!}$
 & \;\; foreach $c\,{\in}\,\mtV^k$
\end{tabular*}
\end{mdframed}
\end{table}

The representation of CrossCat in this paper assumes that data within a
cluster is sampled jointly (step 3), marginalizing over
cluster-specific distributional parameters:
\begin{align*}
M_c(\x_{[\mtC^k_y,c]}, \lambda_c) =
  \int_{\theta} \prod_{r\in{\mtC^k_y}}
  p(x_{[r,c]}|\theta)p(\theta|\lambda_c) d\theta.
\end{align*}
This assumption suffices for our development of predictive relevance, and is
applicable to a broad class of statistical data types \citep{saad2016} with
conjugate prior-likelihood representations such as %
Beta-Bernoulli for binary, %
Dirichlet-Multinomial for categorical, %
Normal-Inverse-Gamma-Normal for real values, %
and Gamma-Poisson for counts.

Given dataset $\D$, we refer to \citet{obermeyer2014} and \citet{mansinghka2016}
for scalable algorithms for posterior inference in CrossCat, and assume we have
access to an ensemble of $H$ posterior samples
$\set{\hat\bphi^1,\dots,\hat\bphi^H}$ where each $\hat\bphi^h$ is a realization
of all variables in Table~\ref{tab:crosscat-symbols}.

\subsection{Estimating predictive relevance using CrossCat}
\label{subsec:crosscat-estimating}

We now describe how to use posterior samples of CrossCat to efficiently estimate
the predictive relevance probability $\mtR_c(\mtQ, r)$ from %
Eq~\eqref{eq:def-relevance}.
Letting $c$ denote the context variable, we formalize the novel variable $c^*$
as a fresh column in the tabular population which is assigned to the same block
$k$ as $c$ (i.e. $k\,{=}\,v_c\,{=}\,v_{c^*})$.
As shown by \citet{saad2017}, structural dependencies induced by CrossCat's
variable partition are related to an upper-bound on the probability there exists
a statistical dependence between $c$ and $c^*$.
To estimate Eq~\eqref{eq:def-relevance}, we first treat the mutual information
between $X_{[q,c^*]}$ and $X_{[r,c^*]}$ as a derived random variable, which is a
function of their random cluster assignments $z^k_q$ and $z^k_r$,
\begin{align}
(z^k_q, z^k_r) \mapsto
  \mtI(X_{[q,c^*]} : X_{[r,c^*]}|z^k_q, z^k_r,\alpha_1,\lambda_{c^*}).
\end{align}
The key insight, implied by step 3 of the CrossCat prior, is that, conditioned
on their assignments, rows from different clusters are sampled independently,
which gives
\begin{flalign}
&z^k_q \ne z^k_r
  && \notag\\
&{\iff}\,p(x_{[q,c^*]}, x_{[r,c^*]}|z^k_q,z^k_r,\lambda_{c^*},\alpha_1,\D)=
  && \notag\\
&\quad \quad p(x_{[q,c^*]}|z^k_q,\lambda_{c^*},\alpha_1,\D)
  p(x_{[r,c^*]}|z^k_r,\lambda_{c^*},\alpha_1,\D)
  && \notag \\
&{\iff}\,\mtI(X_{[q,c^*]} : X_{[r,c^*]}|z^k_q, z^k_r,\alpha_1,\lambda_{c^*})
  = 0,
\label{eq:cluster-mi}
\end{flalign}
where the final implication follows directly from the definition of mutual
information in Eq~\eqref{eq:def-mi}.
Note that Eq~\eqref{eq:cluster-mi} does not depend on the particular choice of
$\lambda_{c^*}$, and indeed this hyperparameter is never represented explicitly.
Moreover, hyperparameter $\alpha_1$ (corresponding to $\alpha$ in
Figure~\ref{fig:hypothesis}) is the concentration of the Dirichlet process for
CrossCat row partitions.

Eq~\eqref{eq:cluster-mi} implies that we can estimate the probability of non-
zero mutual information between $X_{[r,c^*]}$ and each $X_{[q,c^*]}$ for
$q\,{\in}\,\mtQ$ by forming a Monte Carlo estimate from the ensemble of
posterior CrossCat samples,
\begin{flalign}
&\mtR_c(\mtQ,r) && \notag \\
&= \mathbb{P}\left[
  \bigcap\limits_{q\in\mtQ} \left(
    \mtI(X_{[q,c^*]} : X_{[r,c^*]}) > 0
  \right) \mathrel{\bigg|} \lambda_{c^*}, \alpha_1, \D
  \right] \notag \\
&= \mathbb{P}\left[
  \bigcap\limits_{q\in\mtQ}
    \left(z^{v_c}_q = z^{v_c}_r \right)
    \mathrel{\bigg|} \alpha_1, \D
  \right] \notag \\
&\approx \frac{1}{H}\sum_{h=1}^H \left[
  \mathbb{I}\left[
    \bigcap\limits_{q\in\mtQ} \left(
    \hat{z}^{\hat{v}_c^h,h}_q = \hat{z}^{\hat{v}_c^h,h}_r \right)
    \right]
  \right], \label{eq:crosscat-monte-carlo}
\end{flalign}
where $\hat{v}_c^h$ indexes the context block, and $\hat{z}^{\hat{v}_c^h,h}_r$
denotes cluster assignment of $r$ in the row partition of $\hat{v}_c^h$,
according to the sample $\hat\bphi^h$.
Algorithm~\ref{alg:crosscat-relevance} outlines a procedure (used by the BayesDB
query engine from Figure~\ref{fig:workflow}) for formulating a Monte Carlo based
estimator for a predictive relevance query using CrossCat.

\begin{algorithm}[H]
\footnotesize
\caption{\textsc{\small CrossCat-Predictive-Relevance}}
\label{alg:crosscat-relevance}
\begin{algorithmic}[1]
\algrenewcommand\algorithmicindent{.5em}%
\Require $\left\{
\begin{array}{l}
\mbox{CrossCat samples:}\; \hat\bphi^h\; \mbox{for}\; h=1{,}\dots{,}H\\
\mbox{query rows:}\; \mtQ = \set{{q_i} : 1\le{i}\le\lvert\mtQ\rvert}\\
\mbox{context variable:}\; c\\
\end{array}
\right.$
\Ensure \mbox{predictive relevance of each existing row in $\D$ to $\mtQ$}\;
\For{$r = 1, \dots, N$}
  \Comment{for each existing row}
  \For{$h = 1, \dots, H$}
  \Comment{for each CrossCat sample}
    \State $k \gets \hat{v}^h_c$
    \Comment{retrieve the context block}
    \For{$q \in \mtQ$}
    \Comment{for each query row}
      \If{$\hat{z}^{k,h}_q \ne \hat{z}^{k,h}_r$}
        \Comment{$r$ and $q$ are different clusters}
        \State $\mtR^h_c(\mtQ, r) \gets 0$
        \Comment{$r$ irrelevant to some $q$}
        \State \textbf{break}
      \EndIf
    \EndFor
    \ElseTwo
      \Comment{$r$ in same cluster as all $q\in\mtQ$}
      \State $\mtR^h_c(\mtQ, r) \gets 1$
      \Comment{$r$ relevant to all $q$}
    \EndElseTwo
  \EndFor
  \State $\mtR_c(\mtQ, r) \gets \frac{1}{H}\sum_{h=1}^{H}\mtR^h_c(\mtQ,r)$
  \Comment{average relevances}
\EndFor
\State \Return $\set{\mtR_c(\mtQ, r): 1\le{r}\le{N}}$
\end{algorithmic}
\end{algorithm}

\subsection{Optimizing the estimator using a sparse matrix-vector
multiplication}
\label{subsec:crosscat-matrix}

In this section, we show how to greatly optimize the naive, nested for-loop
implementation in Algorithm~\ref{alg:crosscat-relevance} by instead computing
predictive relevance for all $r$ through a single matrix-vector multiplication.

Define the pairwise cluster co-occurrence matrix $\mathbf{S}^{k,h}$ for block
$k$ of CrossCat sample $\hat\bphi^h$ to have binary entries %
$\mathbf{S}^{k,h}_{i,j}\,{=}\,\mathbb{I}[%
  \hat{z}^{k,h}_i\,{=}\,\hat{z}^{k,h}_j]$.
Furthermore, let $\mathbf{1}_{\mtQ}$ denote a length-$N$ vector with a 1 at
indexes $q\,{\in}\,\mtQ$ and 0 otherwise.
We vectorize $\mtR_c(\mtQ,r)$ across $r\,{\in}\,[N]$ by:
\begin{align}
\mathbf{u}^h &=
  \frac{1}{\lvert\mtQ\rvert}\; \mathbf{S}^{k,h} \; \mathbf{1}_{\mtQ}
  && h=1,\dots,H
  \label{eq:crosscat-matrix-vector}\\
\mtR_c(\mtQ, \cdot) &=
  \frac{1}{H}\sum_{h=1}^H \mathbf{u}^h.
  \label{eq:crosscat-matrix-vector-average}
\end{align}
The resulting length-$N$ vector $\mathbf{u}^h$ in %
Eq~\eqref{eq:crosscat-matrix-vector} satisfies $\mathbf{u}^h_r\,{=}\,1$ if and
only if $\hat{z}^{k,h}_r\,{=}\,\hat{z}^{k,h}_q$ for all $q\,{\in}\,\mtQ$, which
we identify as the argument of the indicator function in %
Eq~\eqref{eq:crosscat-monte-carlo}.
Finally, by averaging $\mathbf{u}^h$ across the $H$ samples %
in Eq~\eqref{eq:crosscat-matrix-vector-average}, we arrive at the vector of
relevance probabilities.

For large datasets, constructing the $N{\times}{N}$ matrix $\mathbf{S}^{k,h}$
using $\Theta(N^2)$ operations is prohibitively expensive.
Algorithm~\ref{alg:crosscat-pairwise} describes an efficient procedure that
exploits CrossCat's sparsity to build $\mathbf{S}^{k,h}$ in expected time
${\ll}\,O(N^2)$ by using (i) a sparse matrix representation, and (ii) CrossCat's
partition data structures to avoid considering all pairs of rows.
This fast construction means that Eq~\eqref{eq:crosscat-matrix-vector} is
practical to implement for large data tables.

The algorithm's running time depends on %
(i) the number of clusters $\lvert\textsc{\small Set}(\hat\z^k)\rvert$ in
line~\ref{algline:pairwise-num-clusters};
(ii) the average number of rows per cluster $\lvert\hat\mtC^k_y\rvert$ in
line~\ref{algline:pairwise-num-rows};
and (iii) the data structures used to represent $\mathbf{S}^{k,h}$ in
line~\ref{algline:pairwise-data-structs}.
Under the CRP prior, the expected number of clusters is $O(\alpha_1\log(N))$,
which implies an average occupancy of $O(N/(\alpha_1\log(N)))$ rows per cluster.
If the sparse binary matrix is stored with a list-of-lists representation, then
the update in line \ref{algline:pairwise-data-structs} requires $O(1)$ time.
Furthermore, we emphasize that since $\mathbf{S}^{k,h}$ does not depend $\mtQ$,
its cost of construction is amortized over an arbitrary number of queries.

\begin{algorithm}[H]
\footnotesize
\caption{\textsc{\small CrossCat-Co-Occurrence-Matrix}}
\label{alg:crosscat-pairwise}
\begin{algorithmic}[1]
\algrenewcommand\algorithmicindent{.5em}%
\Require{CrossCat sample $\hat\bphi^h$; block index $k$.}
\Ensure Pairwise co-occurrence matrix $\mathbf{S}^{k,h}$
\For{$y \in \textsc{\small Set}(\hat\z^k)$}
\Comment{for each cluster in block $k$}
\label{algline:pairwise-num-clusters}
  \For{$r \in \hat\mtC^k_y$}
    \Comment{for each row in the cluster}
    \label{algline:pairwise-num-rows}
    \State Set $\mathbf{S}^{k,h}_{r,j} =1$, where $j\in\hat\mtC^k_y$
    \Comment{update the matrix}
    \label{algline:pairwise-data-structs}
  \EndFor
\EndFor
\State \Return $\mathbf{S}^{k,h}$
\end{algorithmic}
\end{algorithm}

\subsection{Computing predictive relevance probabilities for query records that
are not in the database}
\label{subsec:crosscat-appending}

We have so far assumed that the query records must consist of items that already
exist in the database.
This section relaxes this restrictive assumption by illustrating how to compute
relevance probabilities for search records which do not exist in $\D$, and are
instead specified by the user on a per-query basis (refer to the BQL query in
Figure~\ref{fig:workflow} for an example of a hypothetical query record).
The key idea is to (i) incorporate the new records into each CrossCat sample
$\hat\bphi^h$ by using a Gibbs-step to sample cluster assignments from
the joint posterior \citep{neal2000}; (ii) compute %
Eq~\eqref{eq:crosscat-matrix-vector} on the updated samples; and (iii)
unincorporate the records, leaving the original samples unmutated.

Letting $\set{\x_{[N+i]}: 1\le{i}\le{t}}$ denote $t$ (partially observed) new
rows and $\mtQ=\set{N{+}1,\dots,N{+}t}$ the query, we compute $\mtR_c(\mtQ, r)$
for all $r$ by first applying
\textsc{\small CrossCat-Incorporate-Record} %
(Algorithm~\ref{alg:crosscat-incorporate}) to each $q\,{\in}\,\mtQ$
sequentially.
Sequential incorporation corresponds to sampling from the sequence of predictive
distributions, which, by exchangeability, ensures that each updated
$\hat\bphi^h$ contains a sample of cluster assignments from the joint
distribution, guaranteeing correctness of the Monte Carlo estimator in %
Eq~\eqref{eq:crosscat-monte-carlo}.
Note that since CrossCat specifies a non-parametric mixture, the proposal
clusters include all existing clusters, plus one singleton cluster
$\max(\z^k)\,{+}\,1$.
We next update the co-occurrence matrices in time linear in the size of the
sampled cluster and then evaluate Eq~\eqref{eq:crosscat-matrix-vector} and
\eqref{eq:crosscat-matrix-vector-average}.
To unincorporate, we reverse
lines~\ref{algline:crosscat-append-cluster}-\ref{algline:crosscat-append-database}
and restore the co-occurrence matrices.
%
%
Figure~\ref{fig:runtime} confirms that the runtime scaling is asymptotically
linear, varying the (i) number of new rows, (ii) fraction of variables
specified for the new rows that are in the context block (i.e. query sparsity),
(iii) number of clusters in the context block, and (iv) number of variables in
the context block.
\begin{algorithm}[H]
\small
\caption{\textsc{\small CrossCat-Incorporate-Record}}
\label{alg:crosscat-incorporate}
\begin{algorithmic}[1]
\Require{CrossCat sample $\bphi$; context $c$; new row $\x_{N+1}$}
\Ensure Updated crosscat sample $\bphi'$
\State $k \gets v_c$ \Comment{Retrieve block of context variable}
  \label{algline:crosscat-incorporate-block}
\State $Y \gets \max(\z^k)+1$ \Comment{Retrieve proposal clusters}
\For{$y\,{=}\,1,\dots,Y$} \Comment{Compute cluster probabilities}
  \State $n_y \gets \begin{cases}
    \left\vert{\mtC^k_y}\right\vert & \text{if } y \in \z^k \\
    \alpha_1 & \text{if } y = \max({\z^k})\,{+}\,1
    \end{cases}$
  \State $l_y \gets
    \left(
      \prod_{c\in\mtV^k} M_c(x_{[N{+}1,c]}|\x_{[\mtC^k_y,c]}, \lambda_c)
    \right)n_y$
\EndFor
\State $z^k_{N+1} \sim \textsc{Categorical}(l_1,\dots,l_{Y})$
  \Comment{Sample cluster}
\State ${\z'}^k \gets \z^k \cup \set{z^k_{N+1}}$
  \Comment{Append cluster assignment}
  \label{algline:crosscat-append-cluster}
\State ${\mtC'}^k_{z^k_{N{+}1}} \gets \mtC^k_{z^k_{N+1}} \cup \set{N{+}1}$
  \Comment{Append row to cluster}
  \label{algline:crosscat-append-row}
\State $\D' \gets \D \cup \set{\x_{[N+1,\mtV^k]}}$
    \Comment{Append record to database}
    \label{algline:crosscat-append-database}
\State \Return $\bphi'$
    \Comment{Return the updated sample}
\end{algorithmic}
\end{algorithm}

\input{figures/runtime}

\input{figures/gapminder}

\input{figures/usa}

\input{figures/humans}

%% file: figures/runtime.tex

\begin{figure}[ht]
\centering
\includegraphics[width=\linewidth]{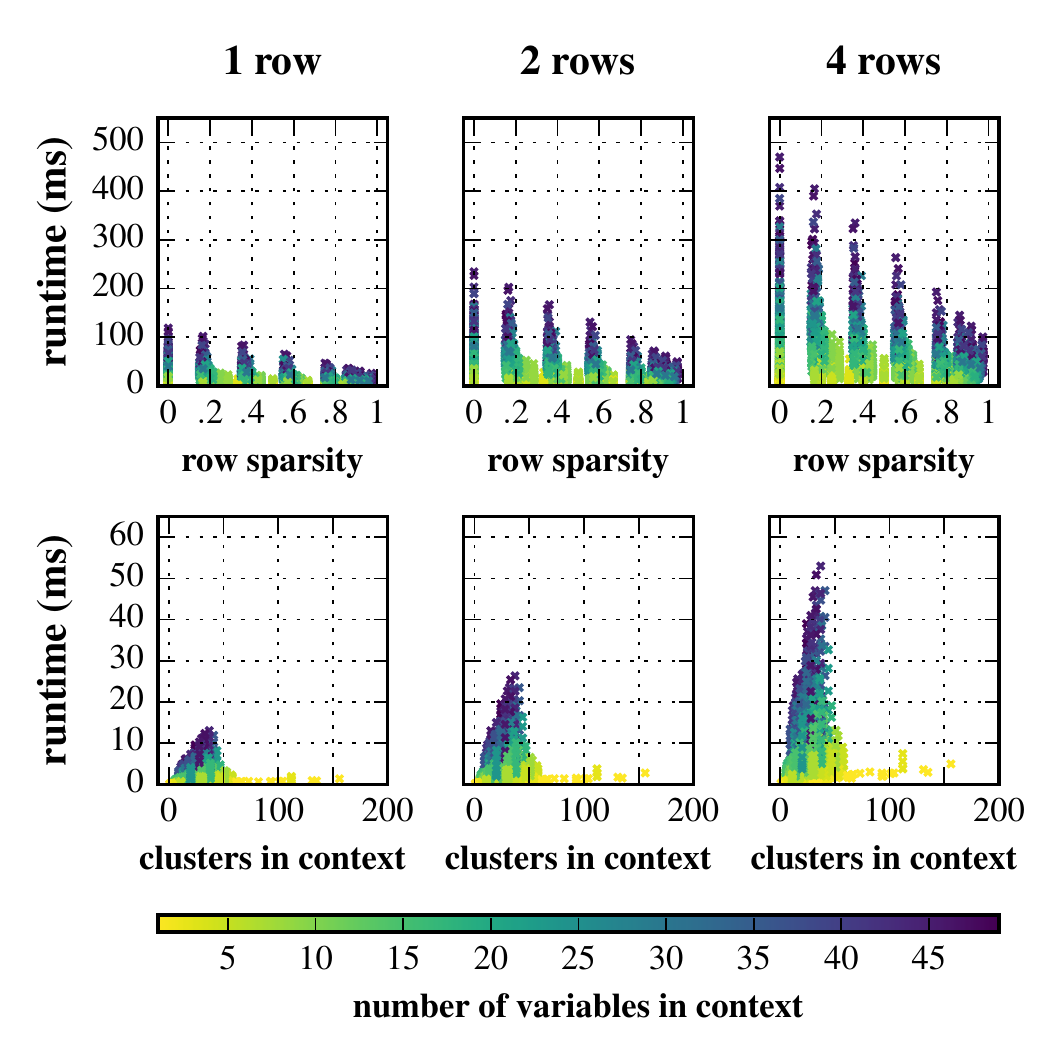}
\caption{
Empirical measurements of the asymptotic scaling of
\textsc{\small CrossCat-Incorporate-Record} %
(Algorithm~\ref{alg:crosscat-incorporate}) on the Gapminder dataset
(Section~\ref{sec:applications}).
The color of each measurement indicates the number of variables in the
block of the context variable; each column shows a different number of records
(1, 2, 4, and 8) incorporated by the algorithm.
The top panels shows that, for a fixed number of variables in the context, the
runtime (in milliseconds) decays linearly with the sparsity of the hypothetical
records (dimensions which are not in the same block as the context variable are
ignored).
The lower panels show the runtime increasing linearly with the number of
clusters in the context; the number of variables in the context dictates the
slope of the curve.
}
\label{fig:runtime}
\end{figure}

%% file: figures/gapminder.tex

\begin{figure*}[ht]
\begin{subfigure}[b]{.5\linewidth}
\centering
\captionsetup{skip=0pt}
\subcaption*{\underline{Pairwise CrossCat predictive relevances in different contexts}}
    \begin{subfigure}[b]{.5\linewidth}
    \centering
    \includegraphics[width=\linewidth]{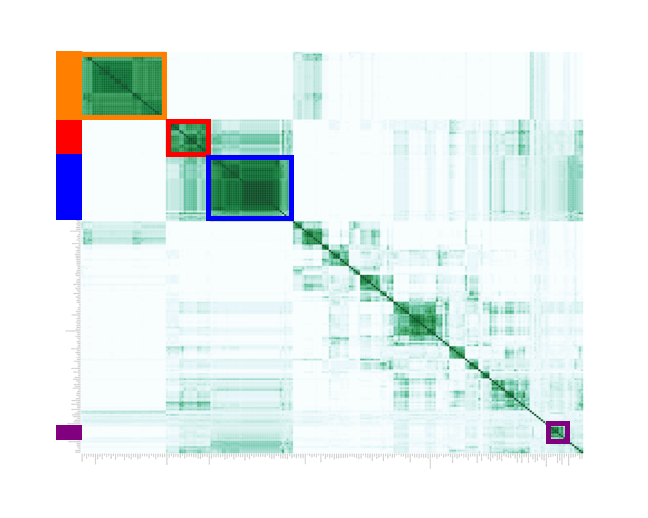}
    \vspace{-.75cm}
    \subcaption{CrossCat (life expectancy)}
    \label{subfig:heatmap-crosscat-lifexp}
    \end{subfigure}%
    \begin{subfigure}[b]{.5\linewidth}
    \centering
    \includegraphics[width=\linewidth]{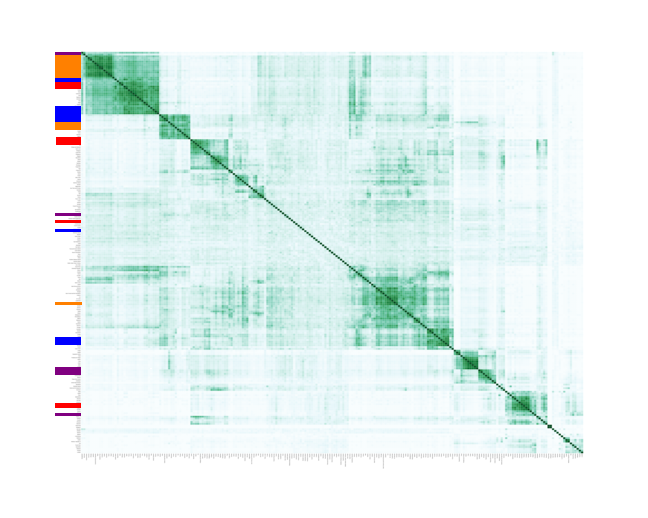}
    \vspace{-.75cm}
    \subcaption{CrossCat (exports, \% gdp)}
    \label{subfig:heatmap-crosscat-exports}
    \end{subfigure}
\end{subfigure}%
\begin{subfigure}[b]{.5\linewidth}
\captionsetup{skip=0pt}
\subcaption*{\underline{Pairwise cosine similarities in different contexts}}
    \begin{subfigure}[b]{.5\linewidth}
    \centering
    \includegraphics[width=\linewidth]{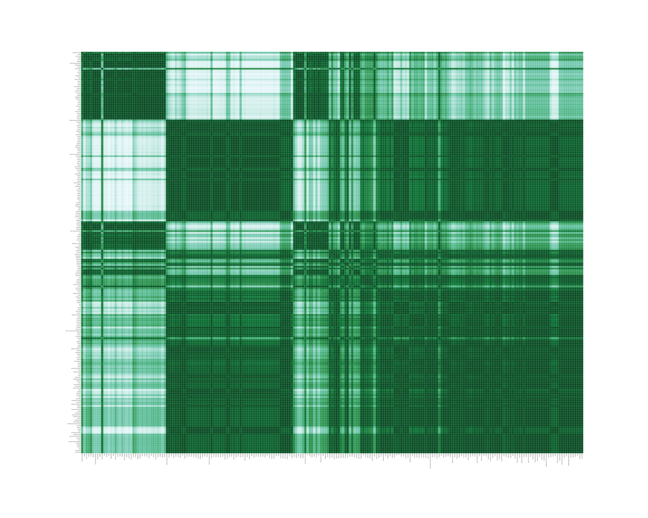}
    \vspace{-.75cm}
    \subcaption{Cosine (life expectancy)}
    \label{subfig:heatmap-cosine-lifexp}
    \end{subfigure}%
    \begin{subfigure}[b]{.5\linewidth}
    \centering
    \includegraphics[width=\linewidth]{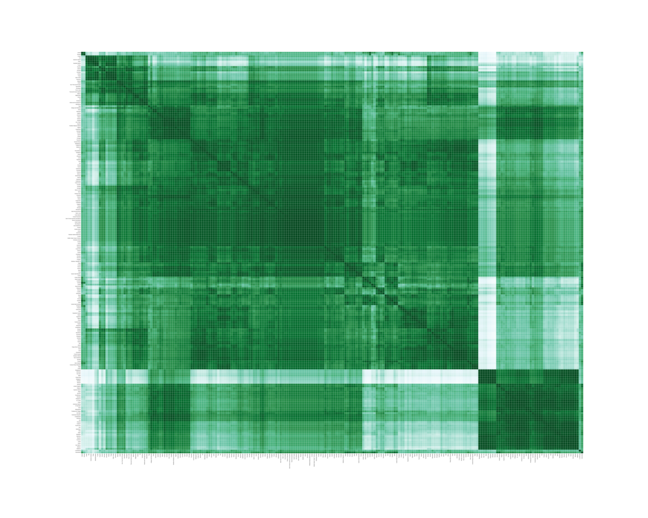}
    \vspace{-.75cm}
    \subcaption{Cosine (exports, \% gdp)}
    \label{subfig:heatmap-cosine-exports}
    \end{subfigure}%
\end{subfigure}

\bigskip

\begin{subtable}[b]{\linewidth}
\footnotesize
\begin{tabularx}{\linewidth}{lX}
\toprule
\textbf{Concept} &
  \textbf{Representative Countries in the Concept}
  \\
\midrule
\textcolor[RGB]{255,128,0}{{\textbf{Low-Income Nations}}}
  & Burundi,
  Ethiopia,
  Uganda,
  Benin,
  Malawi,
  Rwanda,
  Togo,
  Guinea,
  Senegal,
  Afghanistan,
  Malawi
  \\[2pt]
\textcolor[RGB]{255,0,0}{{\textbf{Post-Soviet Nations}}}
  & Russia,
  Ukraine,
  Bulgaria,
  Belarus,
  Slovakia,
  Serbia,
  Croatia,
  Poland,
  Hungary,
  Romania,
  Latvia
  \\[2pt]
\textcolor[RGB]{0,0,255}{{\textbf{Western Democracies}}}
  & France,
  Britain,
  Germany,
  Netherlands,
  Italy,
  Denmark,
  Finland,
  Sweden,
  Norway,
  Australia,
  Japan
  \\[2pt]
\textcolor[RGB]{128,0,128}{{\textbf{Small Wealthy Nations}}}
  & Qatar,
  Bahrain,
  Kuwait,
  Emirates,
  Singapore,
  Israel,
  Gibraltar,
  Bermuda,
  Jersey,
  Cayman Islands
  \\[2pt]
\bottomrule
\end{tabularx}%
\captionsetup{skip=2pt}
\subcaption{
Countries which are mutually predictive in the context of ``life expectancy''
according to CrossCat's relevance matrix
\subref{subfig:heatmap-crosscat-lifexp}.}
\label{subtable:crosscat-clusters}
\end{subtable}

\caption{
\textbf{\subref{subfig:heatmap-crosscat-lifexp} --
\subref{subfig:heatmap-cosine-exports}}
Pairwise heatmaps of countries from the Gapminder dataset in the contexts of
``life expectancy at birth'' and ``exports of goods and services (\% of gdp) '',
using CrossCat predictive relevance and cosine similarity.
Each row and column in a matrix is a country, and a cell value (between 0 and 1)
indicates the strength of match between those two countries.
\textbf{\subref{subtable:crosscat-clusters}} CrossCat learns a sparse set of
relevances; for ``life expectancy'', these broadly correspond to common-sense
taxonomies of countries based on shared geographic, political and macroeconomic
characteristics.
These concepts were manually labeled by inspecting clusters of countries in
matrix \subref{subfig:heatmap-crosscat-lifexp};
the colors in the matrix correspond to countries in the table which belong to
the concept of that color.
Note that the relevance structure differs significantly when ranking in the
context of ``exports, \% gdp'', as shown by the colors in matrix
\subref{subfig:heatmap-crosscat-exports} where the clusters of mutually
relevant countries form a different pattern than in
\subref{subfig:heatmap-crosscat-lifexp}.
Cosine similarity learns dense, noisy sets of spuriously high-ranking
countries with coarser structure, as shown in
\subref{subfig:heatmap-cosine-lifexp} and
\subref{subfig:heatmap-cosine-exports}.
Refer to Appendix~\ref{appx:baselines} for more baselines.
}
\label{fig:gapminder}

\end{figure*}

%% file: figures/usa.tex

\begin{figure}[ht!]

\renewcommand{\ttdefault}{cmtt}
\lstset{
  basicstyle=\ttfamily\footnotesize,
  columns=fullflexible,
  keepspaces=true,
  upquote=true,
  alsoletter={\.,\%},
  morekeywords=[1]{ESTIMATE, FROM, ORDER, BY, RELEVANCE, PROBABILITY,
  EXISTING, ROWS, IN, THE, CONTEXT, OF, TO, AS, WHERE, IS, NOT, DESC, LIMIT},
  keywordstyle=[1]\textcolor{DarkGreen},
  showstringspaces=False,
  stringstyle=\ttfamily\color{Sepia},
  morestring=[b]",
}

\begin{subfigure}{\linewidth}
\begin{lstlisting}
%bql .barplot
...  ESTIMATE "country",
...    RELEVANCE PROBABILITY
...      TO EXISTING ROWS IN
...      ('United States')
...      IN THE CONTEXT OF
...      "life expectancy at birth"
...    AS "rel_us_lifexp"
...  FROM gapminder
...  ORDER BY "rel_us_lifexp" DESC
...  LIMIT 15
\end{lstlisting}
\includegraphics[width=\linewidth]{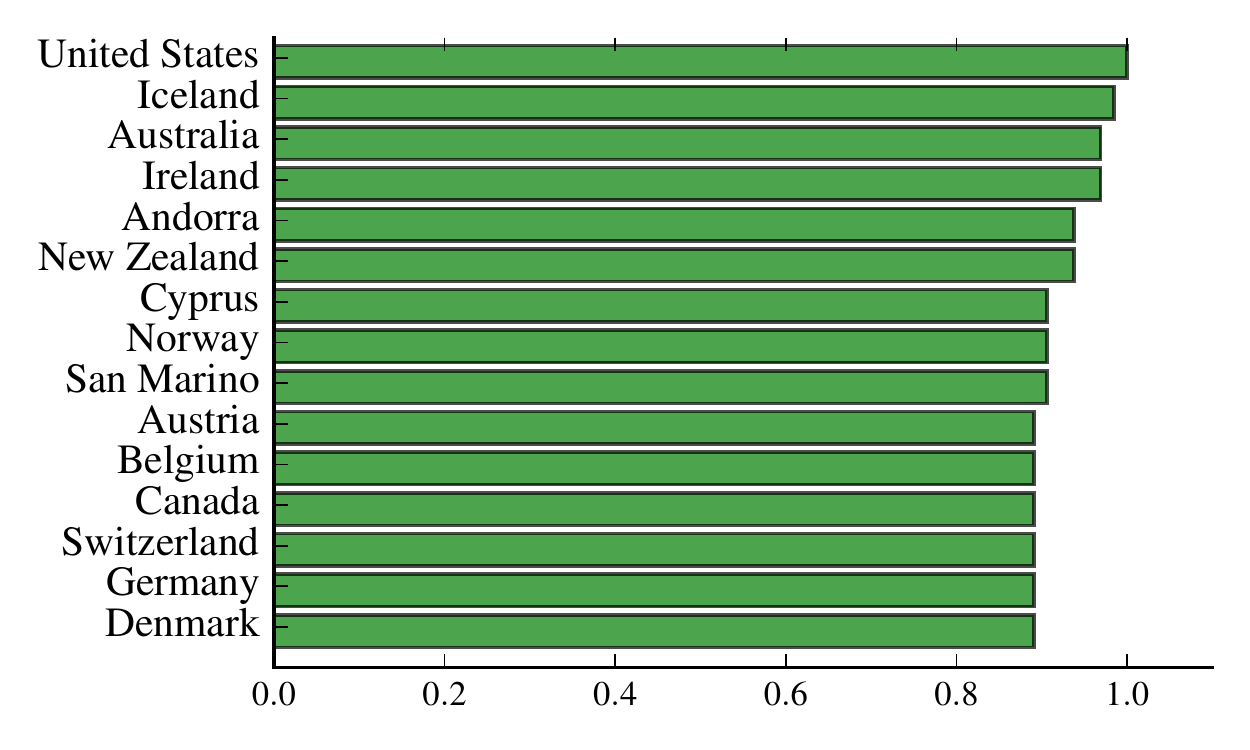}
\captionsetup{skip=-5pt}
\subcaption{\footnotesize Relevance to USA in the context of ``life expectancy''}
\label{subfig:usa-relevant-countries}
\end{subfigure}
\medskip

\begin{subtable}[b]{\linewidth}
\footnotesize
\begin{tabularx}{\linewidth}{|X|}
\hline
Measles, mumps, \& rubella vaccines (\% population)       \\
Under 5 mortality rate                                    \\
Dead children per woman                                   \\
access to improved sanitation facilities (\% population)  \\
access to improved drinking water sources (\% population) \\
human development index                                   \\
body mass index (kg/m2)                                   \\
murder rate (per 100,000)                                 \\
food supply (kilocalories per person)                     \\
contraceptive prevalence (\% women ages 15-49)            \\
alcohol consumption (liters per adult)                    \\
prevalence of tobacco use among adults (\% population)    \\
\hline
\end{tabularx}
\subcaption{Variables in the context of ``life expectancy at birth''}
\label{subfig:usa-context-variables}
\end{subtable}
\caption{\footnotesize
Using BQL to search for the top 15 countries in the Gapminder dataset ranked by
their relevance to the United States in the context of ``life expectancy at
birth'' finds rich, Western democracies with advanced healthcare systems.}
\label{fig:usa}
\end{figure}

%% file: figures/humans.tex

\begin{figure}[ht]
\centering
\includegraphics[width=\linewidth]{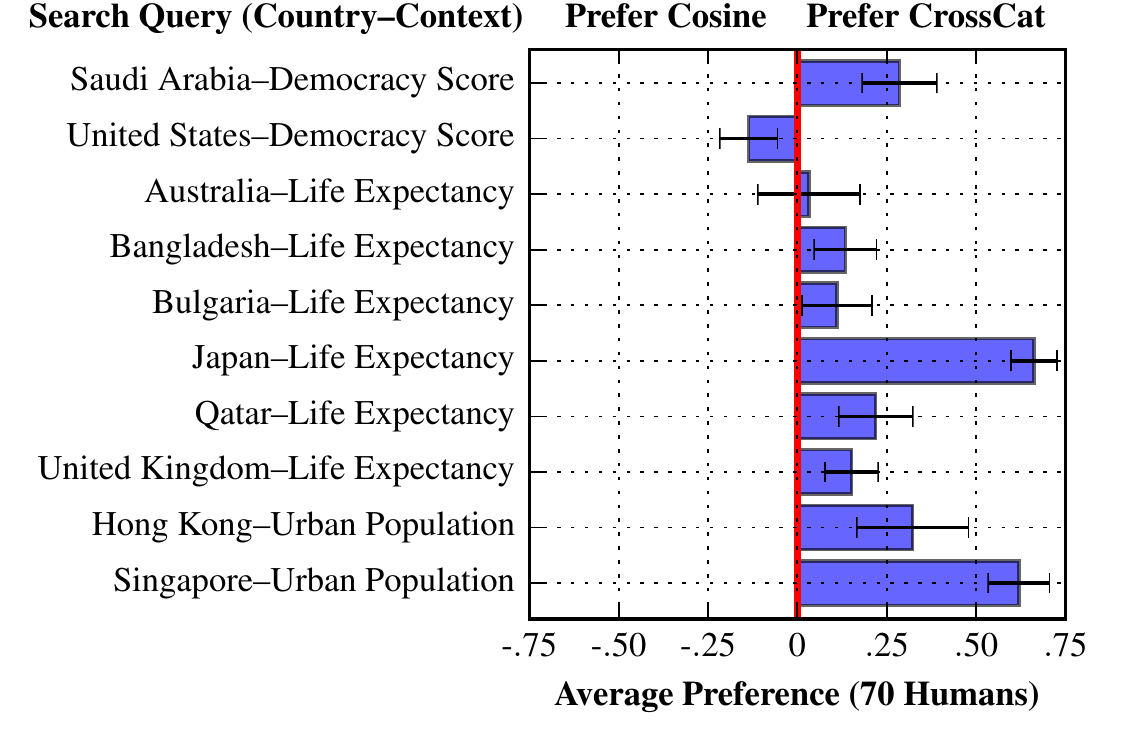}
\captionsetup{skip=0pt}
\caption{\footnotesize
Comparing human preferences for the top-ranked countries returned by cosine
similarity versus CrossCat predictive relevance, in 10 representative search
queries (shown on the y-axis). For each query, human subjects were given the top
10 most relevant countries, according to both cosine and CrossCat, and then
asked to choose which results they preferred, if any. We scored the
responses in the following way: ``countries returned by cosine are more
relevant'' (score = -1); ``countries returned by CrossCat are more relevant''
(score = +1); ``both results are equally relevant'' (score = 0). The x-axis
shows the scores averaged across 70 humans, surveyed on the cloud through
\url{crowdflower.com}. Error bars represent one standard error of the mean. For
most of the queries, human preferences are biased in favor of CrossCat's
rankings. Further details on the experimental design and results are given in
Appendix~\ref{appx:cosine}.}
\label{fig:humans}
\end{figure}

%% file: applications.tex

\section{Applications}
\label{sec:applications}

This section illustrates the efficacy of predictive relevance in BayesDB by
applying the technique to several search problems in real-world, sparse, and
high-dimensional datasets of public interest.%
\footnote{%
Appendix~\ref{appx:cars} contains a further application to a dataset of classic
cars from 1987.
Appendix~\ref{appx:bayesdb} formally describes the integration of
\texttt{RELVANCE PROBABILITY} into BayesDB as an expression in the Bayesian
Query Language (Figure~\ref{fig:workflow}).}

\subsection{College Scorecard}
\label{subsec:applications-scorecard}

The College Scorecard \citep{council2015} is a federal dataset consisting of
over 7000 colleges and 1700 variables, and is used to measure and improve
the performance of US institutions of higher education.
These variables include a broad set of categories such as the campus
characteristics, academic programs, student debt, tuition fees, admission rates,
instructional investments, ethnic distributions, and completion rates.
We analyzed a subset of 2000 schools (four-year institutions) and 100 variables
from the categories listed above.

Suppose a student is interested in attending a city university with a set of
desired specifications.
Starting with a standard SQL Boolean search in %
Figure~\ref{subtable:college-sql} (on p.~\pageref{fig:college}) they find only
one matching record, which requires iteratively rewriting the search conditions
to retrieve more results.

Figure~\ref{subtable:college-bql-1} instead expresses the search query as a
hypothetical row in a BQL \texttt{PREDICTIVE RELEVANCE} query (which invokes the
technique in Section~\ref{subsec:crosscat-appending}).
The top-ranking records contain first-rate schools, but their admission
rates are much too stringent.
In Figure~\ref{subtable:college-bql-2}, the user re-expresses the BQL query to
rank schools by predictive relevance, in the context of instructional
investment, to a subset of the first-rate schools discovered in
\ref{subtable:college-bql-1}.
Combining \texttt{ORDER BY PREDICTIVE RELEVANCE} with Boolean conditions in the
\texttt{WHERE} clause returns another set of top-quality schools with
city-campuses that are less competitive than those in
\ref{subtable:college-bql-1}, but have quantitative metrics that are much better
than national averages.

\subsection{Gapminder}
\label{subsec:applications-gapminder}

Gapminder \citep{rosling2008} is an extensive longitudinal dataset of over
$\sim$320 global macroeconomic variables of population growth, education,
climate, trade, welfare and health for 225 countries.
Our experiments are based on a cross-section of the data from the year 2002.
The data is sparse, with 35\% of the data missing.
Figure~\ref{fig:gapminder} shows heatmaps of the pairwise predictive relevances
for all countries in the dataset under different contexts, and compares the
results to cosine similarity.
Clusters of predictively relevant countries form common-sense taxonomies;
refer to the caption for further discussion.

Figure~\ref{fig:usa} finds the top-15 countries in the dataset ordered by their
predictive relevance to the United States, in the context of ``life expectancy
at birth''.
Table~\ref{subfig:usa-context-variables} shows representative variables which
are in the context; these variables have the highest dependence probability with
the context variable, according a Monte Carlo estimate using 64 posterior
CrossCat samples.
The countries in Figure~\ref{subfig:usa-relevant-countries} are all rich,
Western democracies with highly developed economies and advanced healthcare
systems.

To quantitatively evaluate the quality of top-ranked countries returned by
predictive relevance, we ran the technique on 10 representative search queries
(varying the country and context variable) and obtained the top 10 results for
each query.
Figure~\ref{fig:humans} shows the queries, and human preferences for the results
from predictive relevance versus results from cosine similarity between the
country vectors.
We defined the context for cosine similarity by  the 320-dimensional
vectors down to 10 dimensions and selecting variables which are most dependent
with the context variable according to CrossCat's dependence probabilities.
To deal with sparsity, which cosine similarity cannot handle natively, we
imputed missing values using sample medians; imputation techniques like MICE
\citep{buuren2011} resulted in little difference
(Appendix~\ref{appx:baselines}).

%% file: discussion.tex

\section{Discussion}
\label{sec:discussion}

This paper has shown how to perform probabilistic searches of structured data by
combining ideas from probabilistic programming, information theory, and
nonparametric Bayes.
The demonstrations suggest the technique can be effective on sparse, real-world
databases from multiple domains and produce results that human evaluators often
preferred to a standard baseline.

More empirical evaluation is clearly needed, ideally including tests of hundreds
or thousands of queries, more complex query types, and comparisons with query
results manually provided by human domain experts.
In fact, search via predictive relevance in the context of variables drawn from
learned representations of data could potentially provide a meaningful way to
compare representation learning techniques.
It also may be fruitful to build a distributed implementation suitable for
database representations of web-scale data, including photos, social network
users, and web pages.

Relatively unstructured probabilistic models, such as topic models, proved
sufficient for making unstructured text data far more accessible and useful.
We hope this paper helps illustrate the potential for structured probabilistic
models to improve the accessibility and usefulness of structured data.

%% file: acknowledgements.tex

\subsection*{Acknowledgments}

The authors wish to acknowledge Ryan Rifkin, Anna Comerford, Marie Huber, and
Richard Tibbetts for helpful comments on early drafts. This research was
supported by DARPA (PPAML program, contract number FA8750-14-2-0004), IARPA
(under research contract 2015-15061000003), the Office of Naval Research (under
research contract N000141310333), the Army Research Office (under agreement
number W911NF-13-1-0212), and gifts from Analog Devices and Google.

%% file: bayesdb-integration.tex

\section{Integrating predictive relevance as a ranking function in BayesDB}
\label{appx:bayesdb}

\lstset{
  basicstyle=\ttfamily\footnotesize,
  columns=fullflexible,
  keepspaces=true,
  upquote=true,
  alsoletter={\.,\%},
  morekeywords=[1]{ESTIMATE, FROM, ORDER, BY, PREDICTIVE, RELEVANCE,
  PROBABILITY, EXISTING, ROWS, IN, THE, CONTEXT, OF, TO, AS, WHERE, IS, NOT,
  NEW, ROW, WITH, VALUES, RECORD, HYPOTHETICAL, AND, RECORDS, ASC, DESC,
  AVG},
  keywordstyle=[1]\textcolor{DarkGreen},
  showstringspaces=False,
  stringstyle=\ttfamily\color{Sepia},
  morestring=[b]",
}

This section describes the integration of predictive relevance into BayesDB
\citep{mansinghka2015,saad2016}, a probabilistic programming platform for
probabilistic data analysis.

New syntaxes in the Bayesian Query Language (BQL) allow a user to express
predictive relevance queries where the query set can be an arbitrary combination
of existing and hypothetical records.
We implement predictive relevance in BQL as an expression with the following
syntaxes, depending on the specification of the query records.

\begin{itemize}[leftmargin=*]
\item Query records are existing rows.
\begin{lstlisting}
RELEVANCE PROBABILITY
  TO EXISTING ROWS IN <expression>
  IN THE CONTEXT OF <context-var>
\end{lstlisting}

\item Query records are hypothetical rows.
\begin{lstlisting}
RELEVANCE PROBABILITY
  TO HYPOTHETICAL ROWS WITH VALUES (<values>)
  IN THE CONTEXT OF <context-var>
\end{lstlisting}

\item Query records are existing and hypothetical rows.
\begin{lstlisting}
RELEVANCE PROBABILITY
  TO EXISTING ROWS IN <expression>
  AND HYPOTHETICAL ROWS WITH VALUES (<values>)
  IN THE CONTEXT OF <context-var>
\end{lstlisting}
\end{itemize}

The expression is formally implemented as a 1-row BQL estimand, which specifies
a map $r \mapsto \mtR_c(\mtQ, r)$ for each record in the table.
As shown in the expressions above, query records are specified by the user in
two ways: (i) by giving a collection of \texttt{EXISTING ROWS}, whose primary
key indexes are either specified manually, or retrieved using an arbitrary BQL
\texttt{<expression>}; (ii) by specifying one or more
\texttt{HYPOTHETICAL RECORDS} with their \texttt{<values>} as a list of
column-value pairs.
These new rows are first incorporated using
Algorithm~\ref{alg:crosscat-incorporate} from
Section~\ref{subsec:crosscat-appending} and they are then unincorporated after
the query is finished.
The \texttt{<context-var>} can be any variable in the tabular population.

As a 1-row function in the structured query language, the
\texttt{RELEVANCE PROBABILITY} expression can be used in a variety of settings.
Some typical use-cases are shown in the following examples, where we use only
existing query rows for simplicity.

\begin{itemize}[leftmargin=*]

\item As a column in an \texttt{ESTIMATE} query.
\begin{lstlisting}
ESTIMATE
  "rowid",
  RELEVANCE PROBABILITY
    TO EXISTING ROWS IN <expression>
    IN THE CONTEXT OF <context-var>
  FROM <table>
\end{lstlisting}

\item As a filter in \texttt{WHERE} clause.
\begin{lstlisting}
ESTIMATE
  "rowid"
FROM <table>
WHERE (
    RELEVANCE PROBABILITY
      TO EXISTING ROWS IN <expression>
      IN THE CONTEXT OF <context-var>
    ) > 0.5
\end{lstlisting}

\item As a comparator in an \texttt{ORDER BY} clause.
\begin{lstlisting}
ESTIMATE
  "rowid"
FROM <table>
ORDER BY
    RELEVANCE PROBABILITY
      TO EXISTING ROWS IN <expression>
      IN THE CONTEXT OF <context-var>
[ASC | DESC]
\end{lstlisting}
\end{itemize}

It is also possible to perform arithmetic operations and Boolean comparisons on
relevance probabilities.

\begin{itemize}
\item Finding the mean relevance probability for a set of
\texttt{rowid}s of interest.

\begin{lstlisting}
ESTIMATE
  AVG (
    RELEVANCE PROBABILITY
      TO EXISTING ROWS IN <expression>
      IN THE CONTEXT OF <context-var>
  )
FROM <table>
WHERE "rowid" IN <expression>
\end{lstlisting}

\item Finding rows which are more relevant in some context $c_0$ than in another
context $c_1$.

\begin{lstlisting}
ESTIMATE
  "rowid"
FROM <table>
WHERE (
    RELEVANCE PROBABILITY
      TO EXISTING ROWS IN <expression>
      IN THE CONTEXT OF <context-var-0>
    ) > (
    RELEVANCE PROBABILITY
      TO EXISTING ROWS IN <expression>
      IN THE CONTEXT OF <context-var-1>
    )
\end{lstlisting}
\end{itemize}

%% file: figures/cosine.tex

\begin{figure*}

\section{Predictive relevance and cosine similarity on Gapminder human
evaluation queries}
\label{appx:cosine}
\bigskip

\begin{subtable}{\linewidth}

\begin{subtable}{.2\linewidth}
\centering
\begin{framed}
\subcaption*{\tiny\bfseries Saudi Arabia, Democracy}
\tiny
\begin{tabularx}{.45\linewidth}{c}
\textbf{A} \\
\midrule
\textcolor{red}{Saudi} \\
Oman \\
Libya \\
Kuwait \\
W. Sahara \\
Qatar \\
Bahrain \\
Algeria \\
Iraq \\
Emirates \\
Bhutan \\
\end{tabularx}%
\begin{tabularx}{.45\linewidth}{c}
\textbf{B} \\
\midrule
\textcolor{red}{Saudi} \\
Venezuela \\
Israel \\
Trdad \& Tob \\
Malta \\
Puerto Rico \\
Oman \\
Spain \\
Canada \\
Japan \\
Argentina \\
\end{tabularx}
\includegraphics[width=\linewidth]{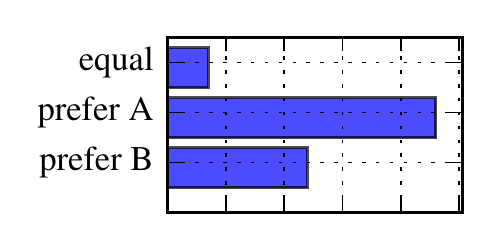}
\end{framed}
\end{subtable}%
\begin{subtable}{.2\linewidth}
\centering
\begin{framed}
\subcaption*{\tiny\bfseries United States, Democracy}
\tiny
\begin{tabularx}{.45\linewidth}{c}
\textbf{A} \\
\midrule
\textcolor{red}{USA} \\
France \\
Finland \\
Norway \\
UK \\
Sweden \\
Estonia \\
Denmark \\
Australia \\
Switzerland \\
Germany \\
\end{tabularx}%
\begin{tabularx}{.45\linewidth}{c}
\textbf{B} \\
\midrule
\textcolor{red}{USA} \\
Australia \\
Ireland \\
Canada \\
UK \\
Iceland \\
Netherlands \\
Austria \\
Denmark \\
Japan \\
New Zealand \\
\end{tabularx}
\includegraphics[width=\linewidth]{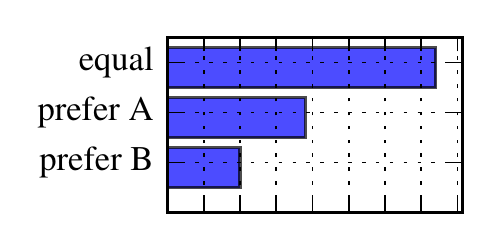}
\end{framed}
\end{subtable}%
\begin{subtable}{.2\linewidth}
\centering
\begin{framed}
\tiny
\subcaption*{\tiny\bfseries Australia, Life Expectancy}
\begin{tabularx}{.45\linewidth}{c}
\textbf{A} \\
\midrule
\textcolor{red}{Australia} \\
Ireland \\
Iceland \\
Andorra \\
United States \\
New Zealand \\
Austria \\
Belgium \\
Canada \\
Switzerland \\
Cyprus \\
\end{tabularx}%
\begin{tabularx}{.45\linewidth}{c}
\textbf{B} \\
\midrule
\textcolor{red}{Australia} \\
Israel \\
Germany \\
Canada \\
Iceland \\
Malta \\
Ireland \\
Finland \\
United States \\
Luxembourg \\
UK \\
\end{tabularx}
\includegraphics[width=\linewidth]{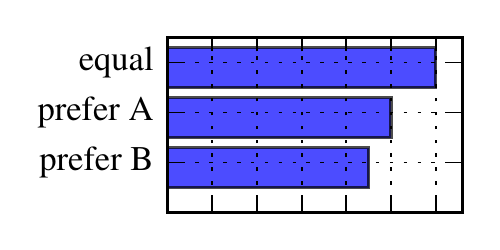}
\end{framed}
\end{subtable}%
\begin{subtable}{.2\linewidth}
\centering
\begin{framed}
\tiny
\subcaption*{\tiny\bfseries Bangladesh, Life Expectancy}
\begin{tabularx}{.45\linewidth}{c}
\textbf{A} \\
\midrule
\textcolor{red}{Bangladesh} \\
Bhutan \\
Papua NG \\
India \\
Gambia \\
Uganda \\
Nepal \\
Timor-Leste \\
Pakistan \\
Mauritania \\
Indonesia \\
\end{tabularx}%
\begin{tabularx}{.45\linewidth}{c}
\textbf{B} \\
\midrule
\textcolor{red}{Bangladesh} \\
India \\
Bhutan \\
Myanmar \\
Indonesia \\
Philippines \\
Nepal \\
Pakistan \\
Mongolia \\
Viet Nam \\
Kyrgyzstan \\
\end{tabularx}
\includegraphics[width=\linewidth]{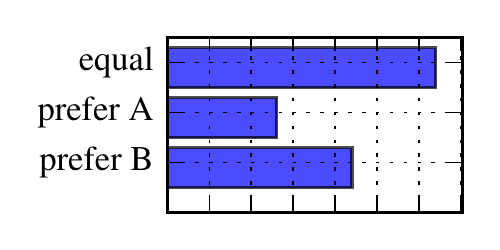}
\end{framed}
\end{subtable}%
\begin{subtable}{.2\linewidth}
\centering
\begin{framed}
\tiny
\subcaption*{\tiny\bfseries Bulgaria, Life Expectancy}
\begin{tabularx}{.45\linewidth}{c}
\textbf{A} \\
\midrule
\textcolor{red}{Bulgaria} \\
Estonia \\
Portugal \\
Macedonia \\
Kuwait \\
Bosnia \\
Hungary \\
Croatia \\
Spain \\
Japan \\
Poland \\
\end{tabularx}%
\begin{tabularx}{.45\linewidth}{c}
\textbf{B} \\
\midrule
\textcolor{red}{Bulgaria} \\
Croatia \\
Poland \\
Serbia \\
Hungary \\
Slovakia \\
Bosnia \\
Belarus \\
Montenegro \\
Estonia \\
Montserrat \\
\end{tabularx}
\includegraphics[width=\linewidth]{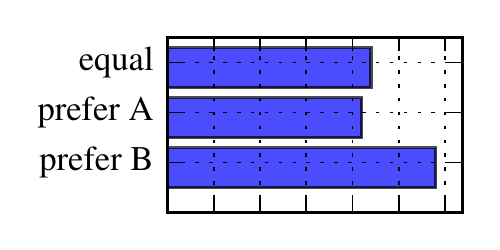}
\end{framed}
\end{subtable}%
\end{subtable}

\bigskip

\begin{subtable}{\linewidth}
\begin{subtable}{.2\linewidth}
\centering
\begin{framed}
\subcaption*{\tiny\bfseries Japan, Life Expectancy}
\tiny
\begin{tabularx}{.45\linewidth}{c}
\textbf{A} \\
\midrule
\textcolor{red}{Japan} \\
Hungary \\
Portugal \\
Spain \\
Slovakia \\
Greece \\
Kuwait \\
Slovenia \\
Emirates \\
Poland \\
Ireland \\
\end{tabularx}%
\begin{tabularx}{.45\linewidth}{c}
\textbf{B} \\
\midrule
\textcolor{red}{Japan} \\
Austria \\
Belgium \\
Canada \\
Switzerland \\
Germany \\
Denmark \\
Finland \\
France \\
UK \\
Netherlands \\
\end{tabularx}
\includegraphics[width=\linewidth]{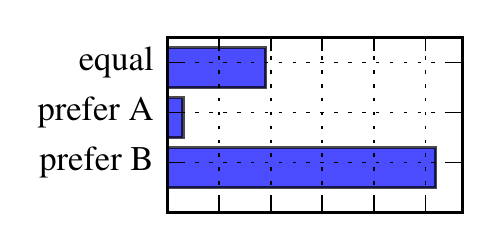}
\end{framed}
\end{subtable}%
\begin{subtable}{.2\linewidth}
\centering
\begin{framed}
\subcaption*{\tiny\bfseries Qatar, Life Expectancy}
\tiny
\begin{tabularx}{.45\linewidth}{c}
\textbf{A} \\
\midrule
\textcolor{red}{Qatar} \\
Emirates \\
Kuwait \\
Bahrain \\
Turks Isld \\
Cayman Isld \\
Guernsey \\
Bermuda \\
Jersey \\
Israel \\
Singapore \\
\end{tabularx}%
\begin{tabularx}{.45\linewidth}{c}
\textbf{B} \\
\midrule
\textcolor{red}{Qatar} \\
Serbia \\
Bosnia \\
Belarus \\
Croatia \\
Montenegro \\
Estonia \\
Bulgaria \\
Lithuania \\
Latvia \\
Saudi Arabia \\
\end{tabularx}
\includegraphics[width=\linewidth]{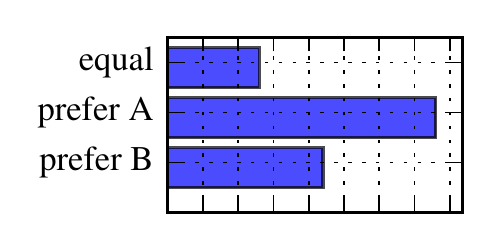}
\end{framed}
\end{subtable}%
\begin{subtable}{.2\linewidth}
\centering
\begin{framed}
\tiny
\subcaption*{\tiny\bfseries UK, Life Expectancy}
\begin{tabularx}{.45\linewidth}{c}
\textbf{A} \\
\midrule
\textcolor{red}{UK} \\
Belgium \\
France \\
Luxembourg \\
Slovenia \\
Germany \\
Malta \\
Canada \\
Finland \\
Ireland \\
Czechia \\
\end{tabularx}%
\begin{tabularx}{.45\linewidth}{c}
\textbf{B} \\
\midrule
\textcolor{red}{UK} \\
Austria \\
Belgium \\
Canada \\
Switzerland \\
Germany \\
Denmark \\
Finland \\
France \\
Japan \\
Netherlands \\
\end{tabularx}
\includegraphics[width=\linewidth]{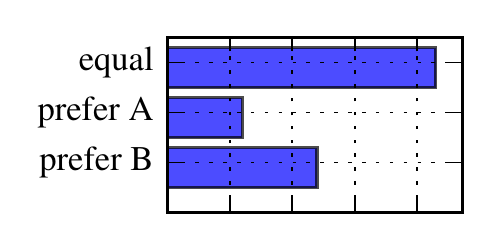}
\end{framed}
\end{subtable}%
\begin{subtable}{.2\linewidth}
\centering
\begin{framed}
\tiny
\subcaption*{\tiny\bfseries Hong Kong, Urban Pop}
\begin{tabularx}{.45\linewidth}{c}
\textbf{A} \\
\midrule
\textcolor{red}{Hong Kong} \\
Italy \\
Mexico \\
Finland \\
Bulgaria \\
Belgium \\
Lithuania \\
Slovakia \\
Poland \\
Lebanon \\
Panama \\
\end{tabularx}%
\begin{tabularx}{.45\linewidth}{c}
\textbf{B} \\
\midrule
\textcolor{red}{Hong Kong} \\
Singapore \\
Austria \\
Canada \\
Greenland \\
Netherlands \\
Andorra \\
Switzerland \\
Ireland \\
Iceland \\
Denmark \\
\end{tabularx}
\includegraphics[width=\linewidth]{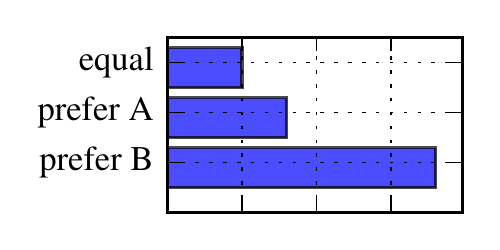}
\end{framed}
\end{subtable}%
\begin{subtable}{.2\linewidth}
\centering
\begin{framed}
\tiny
\subcaption*{\tiny\bfseries Singapore, Urban Pop}
\begin{tabularx}{.45\linewidth}{c}
\textbf{A} \\
\midrule
\textcolor{red}{Singapore} \\
Barbados \\
Oman \\
Norway \\
Romania \\
Libya \\
Algeria \\
Palau \\
Gabon \\
Cuba \\
Switzerland \\
\end{tabularx}%
\begin{tabularx}{.45\linewidth}{c}
\textbf{B} \\
\midrule
\textcolor{red}{Singapore} \\
Hong Kong \\
Gibraltar \\
Andorra \\
Monaco \\
United States \\
San Marino \\
Luxembourg \\
Norway \\
Austria \\
Australia \\
\end{tabularx}
\includegraphics[width=\linewidth]{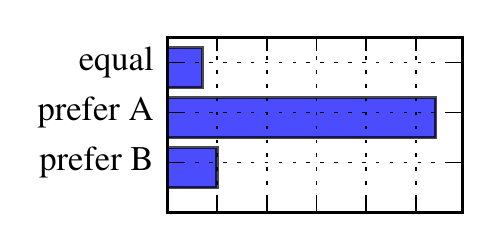}
\end{framed}
\end{subtable}%
\end{subtable}

\caption{%
The top-10 ranking countries returned by predictive relevance and cosine
similarity for each of the 10 queries used for the human evaluation in
Figure~\ref{fig:humans}.
For each country-context search query, we showed seventy subjects (surveyed on
the AI crowdsourcing platform \url{crowdflower.com}) a pair of tables.
We then asked each subject to select the table which contains more relevant
results to the search query, or report that both tables contain equally relevant
results.
The tables above show the top-ranked countries using CrossCat predictive
relevance and cosine similarity, with a histogram of the human responses.
The caption of Figure~\ref{fig:humans} describes how we converted these raw
histograms into scores between -1 and 1 that are displayed in the main text.
The tables showing countries ranked using CrossCat predictive relevance are:
Saudi Arabia (A);
United States (B);
Australia (A);
Bangladesh (B);
Bulgaria (B);
Japan (B);
Qatar (A);
UK (B);
Hong Kong (B);
Singapore (B).
}
\end{figure*}

%% file: figures/baselines.tex

\begin{figure*}

\section{Pairwise heatmaps on Gapminder countries using baseline methods}
\label{appx:baselines}

\begin{subfigure}[b]{\linewidth}
\centering
\captionsetup{skip=0pt}
\subcaption*{\sc Cosine Similarity}
    \begin{subfigure}[b]{.2\linewidth}
    \centering
    \captionsetup{skip=0pt}
    \subcaption*{\scriptsize Median Imputation (5 vars)}
    \includegraphics[width=\linewidth]{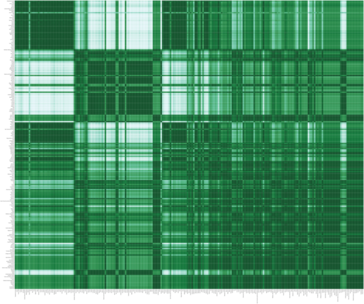}
    \end{subfigure}%
    \begin{subfigure}[b]{.2\linewidth}
    \centering
    \captionsetup{skip=0pt}
    \subcaption*{\scriptsize Median Imputation (10 vars)}
    \includegraphics[width=\linewidth]{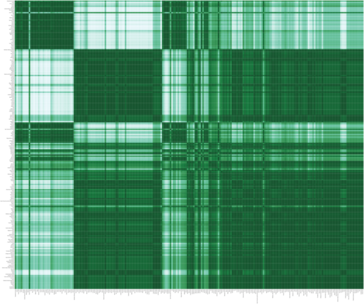}
    \end{subfigure}%
    \begin{subfigure}[b]{.2\linewidth}
    \centering
    \captionsetup{skip=0pt}
    \subcaption*{\scriptsize Median Imputation (15 vars)}
    \includegraphics[width=\linewidth]{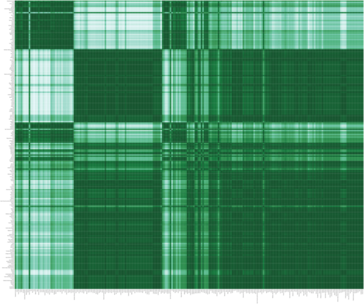}
    \end{subfigure}%
    \begin{subfigure}[b]{.2\linewidth}
    \centering
    \captionsetup{skip=0pt}
    \subcaption*{\scriptsize Median Imputation (20 vars)}
    \includegraphics[width=\linewidth]{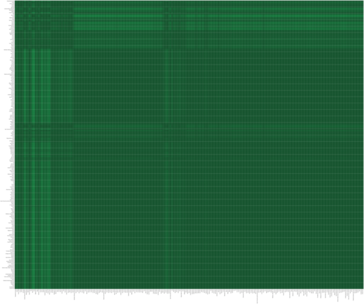}
    \end{subfigure}

    \begin{subfigure}[b]{.2\linewidth}
    \centering
    \captionsetup{skip=0pt}
    \subcaption*{\scriptsize MICE Imputation (5 vars)}
    \includegraphics[width=\linewidth]{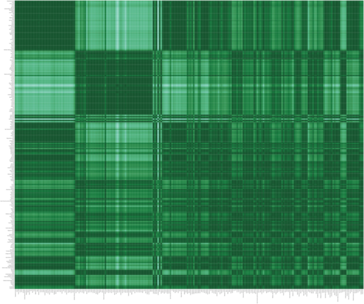}
    \end{subfigure}%
    \begin{subfigure}[b]{.2\linewidth}
    \centering
    \captionsetup{skip=0pt}
    \subcaption*{\scriptsize MICE Imputation (10 vars)}
    \includegraphics[width=\linewidth]{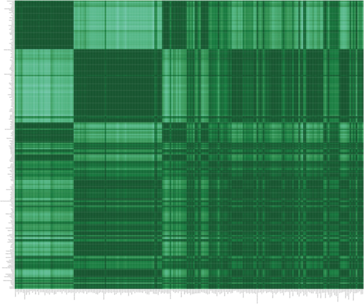}
    \end{subfigure}%
    \begin{subfigure}[b]{.2\linewidth}
    \centering
    \captionsetup{skip=0pt}
    \subcaption*{\scriptsize MICE Imputation (15 vars)}
    \includegraphics[width=\linewidth]{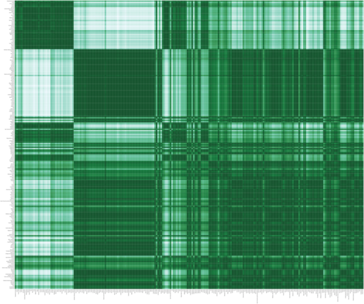}
    \end{subfigure}%
    \begin{subfigure}[b]{.2\linewidth}
    \centering
    \captionsetup{skip=0pt}
    \subcaption*{\scriptsize MICE Imputation (20 vars)}
    \includegraphics[width=\linewidth]{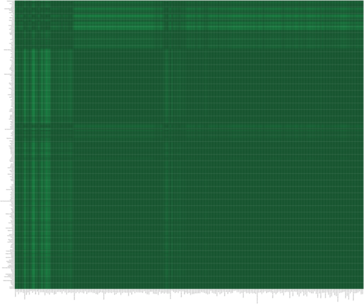}
    \end{subfigure}
\end{subfigure}%

\medskip

\begin{subfigure}[b]{\linewidth}
\centering
\captionsetup{skip=0pt}
\subcaption*{\sc Bray-Curtis Coefficent}
    \begin{subfigure}[b]{.2\linewidth}
    \centering
    \captionsetup{skip=0pt}
    \subcaption*{\scriptsize Median Imputation (5 vars)}
    \includegraphics[width=\linewidth]{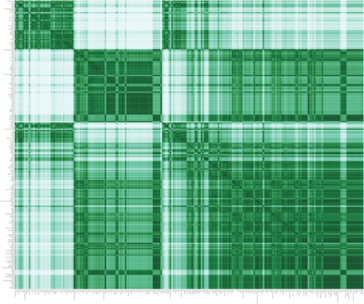}
    \end{subfigure}%
    \begin{subfigure}[b]{.2\linewidth}
    \centering
    \captionsetup{skip=0pt}
    \subcaption*{\scriptsize Median Imputation (10 vars)}
    \includegraphics[width=\linewidth]{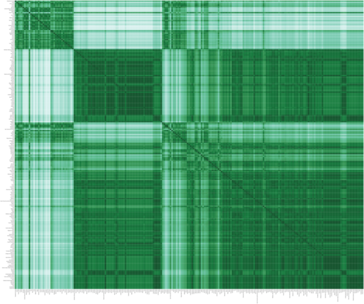}
    \end{subfigure}%
    \begin{subfigure}[b]{.2\linewidth}
    \centering
    \captionsetup{skip=0pt}
    \subcaption*{\scriptsize Median Imputation (15 vars)}
    \includegraphics[width=\linewidth]{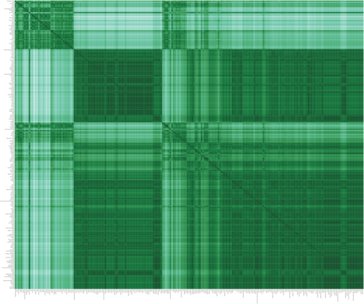}
    \end{subfigure}%
    \begin{subfigure}[b]{.2\linewidth}
    \centering
    \captionsetup{skip=0pt}
    \subcaption*{\scriptsize Median Imputation (20 vars)}
    \includegraphics[width=\linewidth]{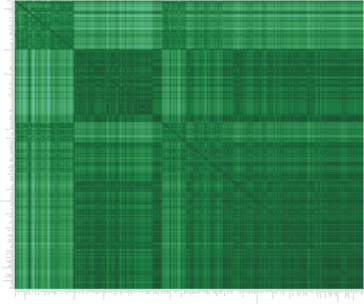}
    \end{subfigure}

    \begin{subfigure}[b]{.2\linewidth}
    \centering
    \captionsetup{skip=0pt}
    \subcaption*{\scriptsize MICE Imputation (5 vars)}
    \includegraphics[width=\linewidth]{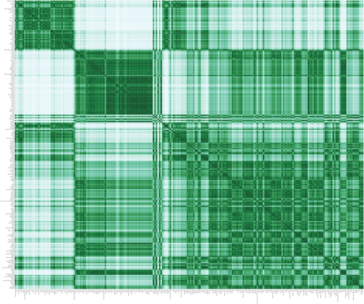}
    \end{subfigure}%
    \begin{subfigure}[b]{.2\linewidth}
    \centering
    \captionsetup{skip=0pt}
    \subcaption*{\scriptsize MICE Imputation (10 vars)}
    \includegraphics[width=\linewidth]{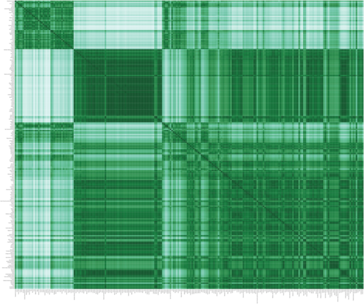}
    \end{subfigure}%
    \begin{subfigure}[b]{.2\linewidth}
    \centering
    \captionsetup{skip=0pt}
    \subcaption*{\scriptsize MICE Imputation (15 vars)}
    \includegraphics[width=\linewidth]{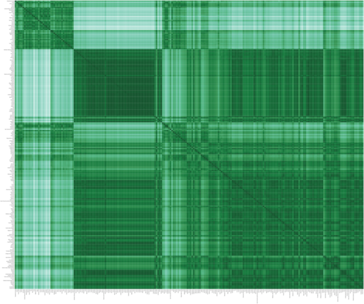}
    \end{subfigure}%
    \begin{subfigure}[b]{.2\linewidth}
    \centering
    \captionsetup{skip=0pt}
    \subcaption*{\scriptsize MICE Imputation (20 vars)}
    \includegraphics[width=\linewidth]{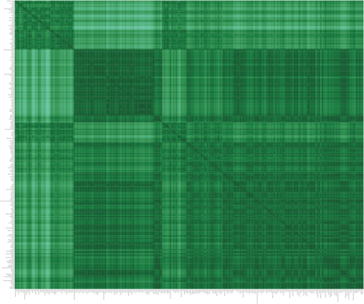}
    \end{subfigure}
\end{subfigure}%

\medskip

\begin{subfigure}[b]{\linewidth}
\centering
\captionsetup{skip=0pt}
\subcaption*{\sc Euclidean Distance}
    \begin{subfigure}[b]{.2\linewidth}
    \centering
    \captionsetup{skip=0pt}
    \subcaption*{\scriptsize Median Imputation (5 vars)}
    \includegraphics[width=\linewidth]{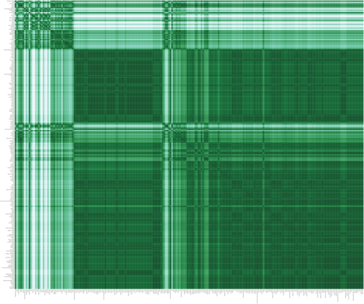}
    \end{subfigure}%
    \begin{subfigure}[b]{.2\linewidth}
    \centering
    \captionsetup{skip=0pt}
    \subcaption*{\scriptsize Median Imputation (10 vars)}
    \includegraphics[width=\linewidth]{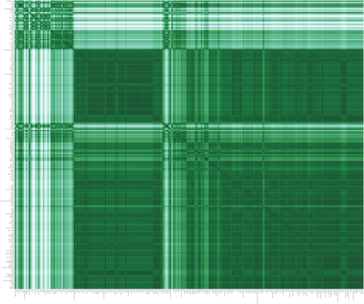}
    \end{subfigure}%
    \begin{subfigure}[b]{.2\linewidth}
    \centering
    \captionsetup{skip=0pt}
    \subcaption*{\scriptsize Median Imputation (15 vars)}
    \includegraphics[width=\linewidth]{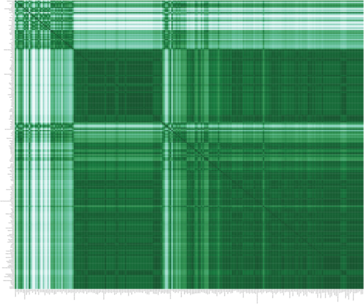}
    \end{subfigure}%
    \begin{subfigure}[b]{.2\linewidth}
    \centering
    \captionsetup{skip=0pt}
    \subcaption*{\scriptsize Median Imputation (20 vars)}
    \includegraphics[width=\linewidth]{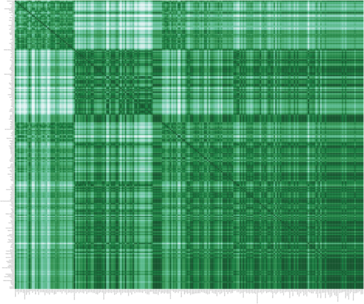}
    \end{subfigure}

    \begin{subfigure}[b]{.2\linewidth}
    \centering
    \captionsetup{skip=0pt}
    \subcaption*{\scriptsize MICE Imputation (5 vars)}
    \includegraphics[width=\linewidth]{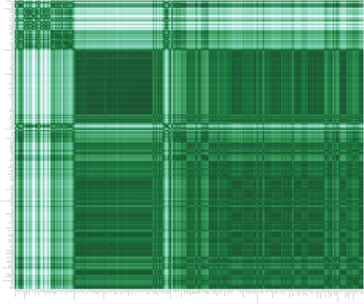}
    \end{subfigure}%
    \begin{subfigure}[b]{.2\linewidth}
    \centering
    \captionsetup{skip=0pt}
    \subcaption*{\scriptsize MICE Imputation (10 vars)}
    \includegraphics[width=\linewidth]{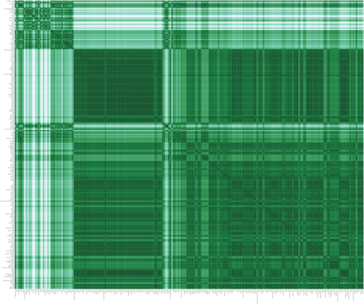}
    \end{subfigure}%
    \begin{subfigure}[b]{.2\linewidth}
    \centering
    \captionsetup{skip=0pt}
    \subcaption*{\scriptsize MICE Imputation (15 vars)}
    \includegraphics[width=\linewidth]{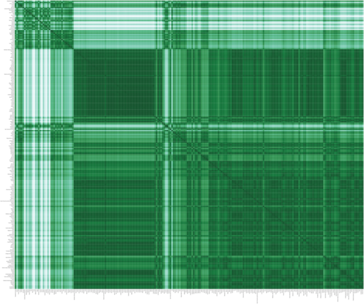}
    \end{subfigure}%
    \begin{subfigure}[b]{.2\linewidth}
    \centering
    \captionsetup{skip=0pt}
    \subcaption*{\scriptsize MICE Imputation (20 vars)}
    \includegraphics[width=\linewidth]{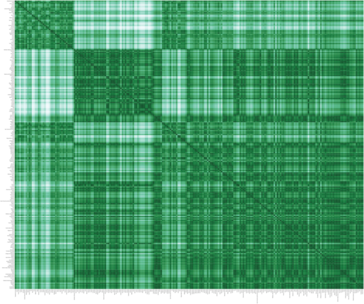}
    \end{subfigure}
\end{subfigure}%

\caption{\footnotesize
Pairwise heatmaps of countries in Gapminder dataset in the context of ``life
expectancy at birth'', using various distance and similarity measures on the
country vectors. %
Each heatmap is labeled with the imputation technique (median or MICE
\citep{buuren2011}), and the number of variables in the context (i.e.
dimensionality of the vectors).
These techniques struggle with sparsity and their structures are much noisier
than the results of relevance probability shown in
Figure~\ref{subfig:heatmap-crosscat-lifexp} and
Table~\ref{subtable:crosscat-clusters}.}

\end{figure*}

%% file: figures/cars.tex

\begin{figure*}
\section{Application to a dataset of 1987 cars}
\label{appx:cars}
\tt\footnotesize

\mdfdefinestyle{codeframe}{
  linecolor=white,%
  leftmargin=0,%
  rightmargin=0,%
  innertopmargin=0,%
  innerbottommargin=0,%
  innerleftmargin=0,%
  innerrightmargin=0,%
}

\renewcommand{\ttdefault}{cmtt}
\lstset{
  basicstyle=\ttfamily\footnotesize,
  columns=fullflexible,
  keepspaces=True,
  upquote=true,
  aboveskip=0pt,
  alsoletter={\.,\%},
  morekeywords=[1]{SELECT, FROM, ORDER, BY, RELEVANCE, PROBABILITY,
  EXISTING, ROWS, IN, THE, CONTEXT, OF, TO, AS, WHERE, IS, NOT, DESC, LIMIT,
  SELECT, AND, HYPOTHETICAL, ROW, CREATE, TABLE, POPULATION, FOR, SCHEMA,
  GUESS, STATTYPES, STATISTICAL, TYPES, FOR, METAMODEL, BASELINE,
  ANALYZE, INITIALIZE, MINUTE, MODELS, DEPENDENCE, PROBABILITY, PAIRWISE,
  VARIABLES},
  keywordstyle=[1]\textcolor{DarkGreen},
  showstringspaces=False,
  stringstyle=\ttfamily\color{Sepia},
  morestring=[b]",
}

\begin{subfigure}[m]{.3\linewidth}
\begin{lstlisting}[aboveskip=\medskipamount]
%bql CREATE TABLE cars_1987_raw
...  FROM 'cars_1987.csv'

%bql SELECT
...   "make",
...   "price",
...   "wheels",
...   "doors",
...   "engine",
...   "horsepower",
...   "body"
...  FROM cars_1987_raw
...  WHERE "price" < 45000
...    AND "wheels" = 'rear'
...    AND "doors" = 'four
...    AND "engine" >= 250
...    AND "horsepower" > 180
...    AND "body" sedan
\end{lstlisting}
\end{subfigure}
\begin{subtable}[m]{.7\linewidth}
\caption{Suppose a customer wishes to purchase a classic car  from 1987 with a
budget of \$45,000 and a desired set of technical specifications. They first
load a csv file of 200 cars with 26 variables into a BayesDB table, and then
specify the search conditions as Boolean filters in a SQL \texttt{WHERE} clause.
Due to sparsity in the table, only one record is returned. To obtain more
relevant results, the user needs to broaden the specifications in the query.}
\label{subfig:cars-sql}
\begin{tabularx}{\linewidth}{Xrllrrr}
\toprule
make     & price & wheels & doors & engine & horsepower & body \\
\midrule
mercedes & 40,960 & rear   & four  & 308    & 184        & sedan \\
\bottomrule
\end{tabularx}
\end{subtable}
\begin{subfigure}[t]{.3\linewidth}
\begin{lstlisting}
%mml CREATE POPULATION
...  cars_1987
...  FOR cars_1987_raw
...  WITH SCHEMA (
...    GUESS STATISTICAL
...    TYPES FOR (*);
...  )

%mml CREATE METAMODEL m FOR cars_1987
...  WITH BASELINE crosscat;

%mml INITIALIZE 100 MODELS FOR m;
%mml ANALYZE m FOR 1 MINUTE;

%bql .heatmap ESTIMATE
...  DEPENDENCE PROBABILITY
...  FROM PAIRWISE VARIABLES
...  OF cars_1987

%bql SELECT
...   "make",
...   "price",
...   "wheels",
...   "doors",
...   "engine-size",
...   "horsepower",
...   "style"
... FROM cars_1987
... ORDER BY
...   RELEVANCE PROBABILITY
...   TO HYPOTHETICAL ROW ((
...    "price" = 42000,
...    "wheels" = 'rear',
...    "doors" = 'four',
...    "engine" = 250,
...    "horsepower" = 180,
...    "body" = 'sedan'
...  ))
...  IN THE CONTEXT OF
...   "price"
... LIMIT 10
\end{lstlisting}
\end{subfigure}%
\begin{subfigure}[t]{.7\linewidth}
\centering
\captionsetup{skip=0pt}
\caption{Building CrosssCat models in BayesDB for the \texttt{cars\_1987}
population learns a full joint probabilistic model over all variables. The
\texttt{ESTIMATE DEPENDENCE PROBABILITY} query allows the user to plot
a heatmap of probable dependencies between car characteristics. The context of
``price'' probably contains the majority of other variables in the search
query.}
\label{subfig:cars-dependence}
\includegraphics[width=.65\linewidth]{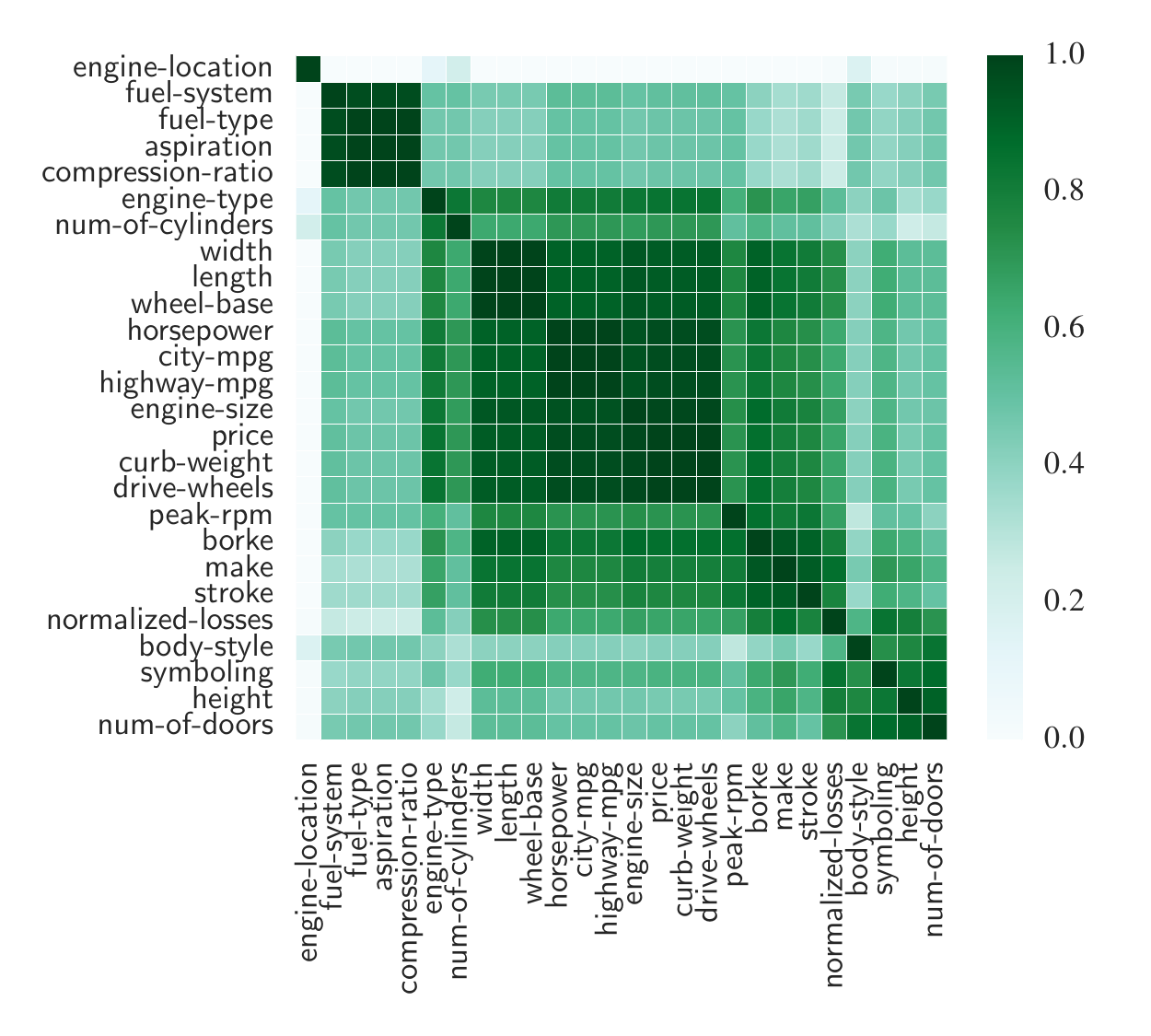}
\captionsetup{skip=10pt}
\subcaption{Using \texttt{ORDER BY RELEVANCE PROBABILITY} in BQL ranks each car
in the table by its relevance to the user's specifications, which are specified
as a hypothetical row. The top-10 ranked cars by probability of relevance to the
search query, in the context of \texttt{price}, are shown below in the table
below. The user can now inspect further characteristics of this subset of cars,
to find ones that they like best.}
\label{subfig:cars-bql}
\begin{tabularx}{\linewidth}{Xrllrrr}
\toprule
 make     & price  & wheels & doors & engine & horsepower & body \\
\midrule
 jaguar   & 35,550 & rear   & four  & 258    & 176        & sedan \\
 jaguar   & 32,250 & rear   & four  & 258    & 176        & sedan \\
 mercedes & 40,960 & rear   & four  & 308    & 184        & sedan \\
 mercedes & 45,400 & rear   & two   & 304    & 184        & hardtop \\
 mercedes & 34,184 & rear   & four  & 234    & 155        & sedan \\
 mercedes & 35,056 & rear   & two   & 234    & 155        & convertible \\
 bmw      & 36,880 & rear   & four  & 209    & 182        & sedan \\
 bmw      & 41,315 & rear   & two   & 209    & 182        & sedan \\
 bmw      & 30,760 & rear   & four  & 209    & 182        & sedan \\
 jaguar   & 36,000 & rear   & two   & 326    & 262        & sedan \\
\bottomrule
\end{tabularx}
\end{subfigure}

\caption{A session in BayesDB for probabilistic model building and search in the
cars dataset \citep{kibler1989}.}
\label{fig:cars}

\end{figure*}